\documentclass[conference]{IEEEtran}
\ifCLASSINFOpdf
  \usepackage[pdftex]{graphicx}
  \graphicspath{{./figures/}}
  \DeclareGraphicsExtensions{.pdf,.png}
\else
\fi
\usepackage{amsmath}
\usepackage{amssymb}
\usepackage{algorithmic}
\usepackage{algorithm}
\ifCLASSOPTIONcompsoc
  \usepackage[caption=false,font=normalsize,labelfont=sf,textfont=sf]{subfig}
\else
  \usepackage[caption=false,font=footnotesize]{subfig}
\fi
\usepackage{url}
\DeclareMathOperator*{\argmax}{arg\,max}
\DeclareMathOperator*{\argmin}{arg\,min}

\begin{document}

\makeatletter
\def\ps@IEEEtitlepagestyle{
  \def\@oddfoot{\mycopyrightnotice}
  \def\@evenfoot{}
}
\def\mycopyrightnotice{
  {\footnotesize
  \begin{minipage}{\textwidth}
  \centering
  Copyright~\copyright~2017 IEEE. Personal use of this material is permitted. Permission from IEEE must be obtained for all other uses, in any current or future media, including reprinting/republishing this material for advertising or promotional purposes, creating new collective works, for resale or redistribution to servers or lists, or reuse of any copyrighted component of this work in other works
  \end{minipage}
  }
}

\title{Large-Scale Stochastic Learning using GPUs}

%\author{\IEEEauthorblockN{Thomas Parnell}
%\IEEEauthorblockA{IBM Research - Zurich\\
%Email: tpa@zurich.ibm.com}
%\and
%\IEEEauthorblockN{Celestine Duenner}
%\IEEEauthorblockA{IBM Research - Zurich\\
%Email: cdu@zurich.ibm.com}
%\and
%\IEEEauthorblockN{Kubilay Atasu}
%\IEEEauthorblockA{IBM Research - Zurich\\
%Email: kat@zurich.ibm.com}
%\and
%\IEEEauthorblockN{Manolis Sifalakis}
%\IEEEauthorblockA{IBM Research - Zurich\\
%Email: emm@zurich.ibm.com}
%\and
%\IEEEauthorblockN{Haris Pozidis}
%\IEEEauthorblockA{IBM Research - Zurich\\
%Email: hap@zurich.ibm.com}}

\author{\IEEEauthorblockN{Thomas Parnell,
Celestine D\"{u}nner,
Kubilay Atasu, 
Manolis Sifalakis and
Haris Pozidis}
\IEEEauthorblockA{IBM Research - Zurich\\
S\"{a}umerstrasse 4, R\"{u}schlikon, Switzerland\\
 Email: \{tpa, cdu, kat, emm, hap\}@zurich.ibm.com}}


\maketitle

\begin{abstract}
In this work we propose an accelerated stochastic learning system for very large-scale applications. Acceleration is achieved by mapping the training algorithm onto massively parallel processors: we demonstrate a parallel, asynchronous GPU implementation of the widely used stochastic coordinate descent/ascent algorithm that can provide up to $35\times$ speed-up over a sequential CPU implementation. In order to train on very large datasets that do not fit inside the memory of a single GPU, we then consider techniques for distributed stochastic learning. We propose a novel method for optimally aggregating model updates from worker nodes when the training data is distributed either by example or by feature. Using this technique, we demonstrate that one can scale out stochastic learning across up to 8 worker nodes without any significant loss of training time. Finally, we combine GPU acceleration with the optimized distributed method to train on a dataset consisting of 200 million training examples and 75 million features. We show by scaling out across 4 GPUs, one can attain a high degree of training accuracy in around 4 seconds: a $20\times$ speed-up in training time compared to a multi-threaded, distributed implementation across 4 CPUs. 
\end{abstract}


% \fi
\IEEEpeerreviewmaketitle

\section{Introduction}

Graphics processing units (GPUs) have found a wide range of applications in machine learning and other scientific fields. Most recently, they have 
become widely adopted to tackle the problem of training deep neural networks \cite{Krizhevsky2012}. Due to their massively parallel architecture, GPUs are well suited to this task, since the training procedure can be naturally expressed as a sequence of matrix operations. By making carefully constructed calls to libraries such as NVIDIA's cuBLAS (or even cuDNN) one can typically achieve the maximum theoretical floating point performance of the processor.

While neural networks are a topic of significant interest, many other machine learning applications rely on other techniques such as fitting generalized linear models for regression or classification \cite[Chapters 3 and 4]{Friedman2001}. While the training of such models can generally also be mapped to a sequence of matrix operations, this tends to apply only when using batch methods such as gradient descent. The batch gradient descent technique updates the model parameters by computing a gradient vector across all available training examples. It is well known that faster convergence can be achieved over batch methods by using stochastic learning algorithms such as stochastic gradient descent (SGD) \cite{Bottou1998} or stochastic coordinate descent (SCD) \cite{Friedman2010}. These algorithms compute an update to the model parameters by computing a gradient using only a single training example or optimizing with respect to a single feature respectively. Compared to batch methods, such algorithms are inherently sequential and each successive iteration involves only a single vector inner product computation. Furthermore, these vectors are sparse for many datasets of interest. For these reasons, mapping such algorithms onto GPUs become more difficult since, within a single iteration, one cannot benefit significantly from the massively parallel architecture. 

A further challenge is the issue of limited GPU memory. State-of-the-art GPUs have a main memory capacity of up to 16GB. Modern internet-scale datasets \cite{criteo2015} can grow many times larger than this. Therefore, stochastic learning algorithms that run on a single GPU are of limited interest and to build a useful GPU-based implementation it is necessary to consider distributed techniques. Distribution of stochastic learning is an active field of research and many promising techniques have been proposed. In \cite{Li2014} a method was proposed whereby worker nodes perform stochastic updates of a local model and asynchronously communicate their model updates to a parameter server. Alternatively, one may consider synchronous techniques such as \cite{Jaggi2014} in which worker nodes perform stochastic updates using the data that is locally available to them and after a number of steps, all updates are aggregated on a master node and the resulting model (or some representation thereof) is then broadcast back to the workers for the next round. Unless the data has been partitioned in a way to exploit underlying structure, distributed algorithms tend to converge slower (in terms of number of model updates) compared to their non-distributed counterparts due to the delay incurred in sharing model updates between workers \cite{Bekkerman2011}. However, when one is truly dealing with very large data, scaling out across multiple machines (or indeed GPUs) becomes a necessity rather than a choice.

In this work we will begin with an overview of the ridge regression problem ($L^2$-regularized linear regression) and explain how it can be solved using stochastic coordinate descent/ascent techniques in its primal formulation as well as its dual formulation. While we have opted to focus on ridge regression for the sake of simplicity, stochastic coordinate methods are used in the field of machine learning to solve other problem such as regression with elastic net regularization as well as support vector machines. We proceed to review the state-of-the-art implementations of SCD on the CPU, covering both single-threaded and multi-threaded implementations. We will then present a twice parallel, asynchronous implementation of SCD (TPA-SCD) that is specifically designed to take advantage of the massively parallel hardware that is available on modern GPUs. We will demonstrate that this new implementation can achieve up to a $35\times$ speed-up in training time relative to existing single-threaded algorithms. We will then turn our attention to the problem of training on very large datasets and discuss how SCD can be distributed across multiple workers nodes that communicate over a network. We will consider the case where the training data is distributed across the workers nodes by training example as well as the case where the data is distributed by feature. We will then study a core component of synchronous distributed learning algorithms: the aggregation step. We propose a novel technique for direct optimization of the aggregation step for the case of distributed ridge regression. We will demonstrate that by using optimized aggregation, it is possible to scale out across up to 8 nodes without experiencing any significant slow-down in training time. Finally, we will combine these two techniques and demonstrate distributed stochastic learning across clusters of GPUs. We show that by using a cluster of 4 GPUs running TPA-SCD it is possible to train a 40GB sample of the criteo dataset (200 million examples, 75 million non-zero features) to a high degree of accuracy in around 4 seconds, providing a $20\times$ speed-up over a state-of-the-art multi-threaded, distributed implementation. 

\section{Ridge Regression}
Let $A\in\mathbb{R}^{N\times M}$ denote the training data matrix where $N$ is the number of training examples 
and $M$ is the number of features. The corresponding labels for the training examples are provided by the vector 
$y\in\mathbb{R}^N$. The parameter $\lambda\in\mathbb{R}^+$ controls regularization and prevents over-fitting. The operator $||.||$ denotes the standard $L^2$-norm and the operator $\left<.\right>$ denotes the Euclidean inner product between two vectors.

\subsection{Primal Form}
Ridge regression in its primal form is defined by the following objective function:
\begin{eqnarray}
	\mathcal{P}(\beta) = \frac{1}{2N}||A\beta-y||^2 + \frac{\lambda}{2}||\beta||^2,\label{eq:primal_obj}
\end{eqnarray}
where $\beta\in\mathbb{R}^M$ are the primal model weights. The function $\mathcal{P}(\beta)$ is strongly convex 
and the optimal values are given by:
\begin{eqnarray}
	\beta^* = \argmin_{\beta}\mathcal{P}(\beta).\nonumber
\end{eqnarray}
Coordinate descent methods aim to iteratively minimize the primal objective (\ref{eq:primal_obj}) by successively optimizing individual coordinates.
The partial derivative of the primal function with respect to the $m$-th coordinate is given by:
\begin{eqnarray}
	\frac{\partial P(\beta)}{\partial\beta_m} = \frac{1}{N}\left<A\beta-y,a_m\right> + \lambda\beta_m,\nonumber
\end{eqnarray}
where $a_m$ denotes the $m$-th column of data matrix $A$. 
Let $\beta^{(t)}$ denote the approximate solution at iteration $t$. By following the approach of \cite{Friedman2010} and optimizing 
with respect to the $m$-th coordinate while keeping the model weights for all other coordinates fixed, one obtains the following update rule:
\begin{eqnarray}
	\beta^{(t+1)} = \beta^{(t)} + \left(\frac{\left<y-w^{(t)},a_m\right>-\lambda N\beta_m^{(t)}}{||a_m||^2 + \lambda N}\right)e_m,\label{eq:update_primal}
\end{eqnarray}
where $w^{(t)}=A\beta^{(t)}\in\mathbb{R}^N$ is known as the \textit{shared vector} at iteration $t$ and $e_m\in\{0,1\}^M$ is a vector that is all-zero apart from the $m$-th coordinate. The following 
update rule for the shared vector follows easily:
\begin{eqnarray}
	w^{(t+1)} = w^{(t)} + a_m(\beta^{(t+1)}_m-\beta^{(t)}_m).\nonumber
\end{eqnarray}

%% -----
\subsection{Dual Form}
Ridge regression in its dual form is defined by the following objective function:
\begin{eqnarray}
	\mathcal{D}(\alpha) = -\frac{N}{2}||\alpha||^2 - \frac{1}{2\lambda}||A^T\alpha||^2 + \alpha^T y,\label{eq:dual_obj}
\end{eqnarray}
where $\alpha\in\mathbb{R}^N$ are the dual model weights. The optimal value for the model weights can be found by 
solving the following maximization problem:
\begin{eqnarray}
	\alpha^* = \argmax_{\alpha}\mathcal{D}(\alpha).\nonumber
\end{eqnarray}
The dual problem can also be solved using iterative techniques that maximize the objective function (\ref{eq:dual_obj}) using a single coordinate at a time. The partial derivative with respect to the $n$-th coordinate is given by:
\begin{eqnarray}
	\frac{\partial D(\alpha)}{\partial\alpha_n} = -N\alpha_n - \frac{1}{\lambda}\left<A^T\alpha, \bar{a}_n\right> + y_n,\nonumber
\end{eqnarray}
where $\bar{a}_n$ denotes the $n$-th row of the data matrix $A$. Let $\alpha_i^{(t)}$ denote the estimate of the optimal model weights at iteration $t$. It was shown in \cite{Shalev2013} that one can optimize the dual objective function for a selected coordinate $n$ while keeping the model weights for all other coordinates fixed, leading to the following update rule:
\begin{eqnarray}
	\alpha^{(t+1)} = \alpha^{(t)} + \left(\frac{\lambda y_n - \left<\bar{w}^{(t)},\bar{a}_n\right> - \lambda N\alpha_i^{(t)} }{\lambda N + ||\bar{a}_n||^2}\right)\bar{e}_n,\label{eq:update_dual}
\end{eqnarray}
where $\bar{w}^{(t)}=A^T\alpha^{(t)}\in\mathbb{R}^M$ denotes the dual shared vector and $\bar{e}_n\in\{0,1\}^N$ is a vector that is all-zero apart from the $n$-th coordinate. The update rule for the dual shared vector is given by:
\begin{eqnarray}
	\bar{w}^{(t+1)} = \bar{w}^{(t)} + \bar{a}_n(\alpha^{(t+1)}_n-\alpha^{(t)}_n).\nonumber
\end{eqnarray}
\subsection{Duality Gap}
The Fenchel-Rockafellar duality theorem \cite{Ekeland1999} dictates that $\mathcal{P}(\beta^*)=\mathcal{D}(\alpha^*)$ and the following conditions must hold for the optimal values of $\beta^*$ and $\alpha^*$:
\begin{eqnarray}
\beta^* &=& \frac{1}{\lambda}A^T\alpha^*,\label{eq:duality-1}\\
\alpha^* &=& \frac{1}{N}(y-A\beta^*).\label{eq:duality-2}
\end{eqnarray}
In order to compare the convergence behavior of the two methods we can define the duality gap for the primal, dual algorithms respectively as follows:
\begin{eqnarray}
G_{\mathcal{P}}\left(\beta^{(t)}\right) &=& \left|\mathcal{P}\left(\beta^{(t)}\right)-\mathcal{D}\left(\frac{1}{N}\left(y-A\beta^{(t)}\right)\right)\right|,\nonumber \\
G_{\mathcal{D}}\left(\alpha^{(t)}\right) &=& \left|\mathcal{P}\left(\frac{1}{\lambda}A^T\alpha^{(t)}\right)-\mathcal{D}\left(\alpha^{(t)}\right)\right| \nonumber.
\end{eqnarray}
We use the duality gap to compare convergence of algorithms that solve the primal and dual formulations of ridge regression since it does not depend on the scale of either objective and its limit when the number of iterations is large is known: it must always converge to zero. In the next section we will implement these algorithms and compare their convergence behavior.

\section{Stochastic Learning on the GPU}
\label{sec:gpu}
In this section we will consider how to implement stochastic coordinate descent methods. Since the dual maximization problem can always be transformed into a minimization by applying a change of sign, in what follows we will solely refer to stochastic coordinate descent (SCD) methods for solving both formulations of ridge regression. We first review the standard sequential implementation on the CPU and the existing work on asynchronous, multi-threaded implementations. We will then describe a new implementation (TPA-SCD) that is designed to exploit the massively parallel computing resources provided by modern GPUs. 

\subsection{Sequential CPU Implementation}

\begin{algorithm}
\caption{Sequential SCD \cite{Friedman2010}.}
\label{alg:rcd_sync}
\begin{algorithmic}
\STATE{Initialize: $t=0,\beta^{(t)}=0,w^{(t)}=0$.}
\FOR{$epoch=1\ldots,n_{epochs}$}
\STATE{Generate random permutation of features $P_{epoch}$.}
\FOR{$j=1\ldots M$}
\STATE{Update randomized coordinate:}
\STATE{$m = P_{epoch}(j)$}
\STATE{$\Delta\beta = \left(||a_m||^2 + N\lambda\right)^{-1}\left(\left<y-w^{(t)},a_m\right> - N\lambda\beta^{(t)}_m\right)$}
\STATE{$\beta^{(t+1)}=\beta^{(t)}+e_m\Delta\beta$}
\STATE{Update shared vector:}
\STATE{$w^{(t+1)} = w^{(t)} + a_m\Delta\beta$}
\STATE{$t = t + 1$}
\ENDFOR
\ENDFOR
\end{algorithmic}
\end{algorithm}

In Algorithm \ref{alg:rcd_sync} we review the algorithm proposed in \cite{Friedman2010} for sequential SCD. The algorithm is presented for the primal form of ridge regression, however the equivalent variation for the dual formulation is almost identical up to the update rule (\ref{eq:update_dual}). For the primal form of the algorithm, one epoch is defined to be one pass through all the $M$ permuted features. Conversely, for the dual form an epoch is defined as one pass through all the $N$ permuted training samples. Both variants of the algorithm have been implemented in C++. Optimization flags were passed to the gcc compiler to ensure that vectorization occurs when evaluating the inner products. The data matrix and model parameters are represented using $32$-bit floating point data types. The implementation has been designed assuming that the data matrix $A$ is sparse, so that any unnecessary computation is avoided. 

\subsection{Asynchronous CPU Implementations}

SCD is an inherently sequential algorithm and is thus non-trivial to parallelize. However, recent work into asynchronous techniques has shown that it is possible to accelerate stochastic learning algorithms by running multiple threads (each updating using a single coordinate or example) that read the current value of the model parameters (and any associated vectors) from shared memory and write back their updates without using complex locking schemes such as those proposed in \cite{Kaleem2015}. In \cite{Recht2011} an asynchronous implementation of stochastic gradient descent was proposed (``Hogwild!") that comprises many parallel threads each computing the gradient using a random training example and updating the model weights using atomic operations. While this work significantly developed the concept of asynchronous learning, asynchronous coordinate descent methods were not considered.

In \cite{Tran2015}, an asynchronous version of Algorithm \ref{alg:rcd_sync} was proposed (A-SCD) whereby the inner loop over the shuffled coordinates is parallelized across multiple CPU threads. Since different threads can potentially write updates to the shared vector in the same location, an atomic addition was used to ensure that updates to the shared vector are always applied. The authors found that, while a good speed-up was attained, issues arose due to the shared vector becoming inconsistent with the model weights. To resolve this problem, a scheme for occasionally re-computing the shared vector was proposed. In \cite{Hsieh2015} it was reported that one can achieve faster convergence if instead of using atomic addition, one allows a ``wild" behavior where updates to the shared vector can be overwritten or not applied at all.  While the authors reported an almost linear speed-up in training time using this algorithm (PASSCoDe-Wild), it was shown that such an implementation will converge to a solution that violates the optimality conditions (\ref{eq:duality-1}) and (\ref{eq:duality-2}). 

Finally, in \cite{Liu2015}, an asynchronous coordinate descent algorithm was proposed (AsySCD) and close-to-linear scaling was demonstrated using a 40-core CPU. 
This algorithm differs from Algorithm \ref{alg:rcd_sync} in two important respects.
Firstly, instead of optimizing for each coordinate exactly, a small gradient descent step is taken thus introducing an additional step size parameter that must be tuned. 
Secondly, the algorithm is implemented without the use of a shared vector. Instead, the computation of a Hessian matrix is required. 
This takes a considerable amount of time and significantly increases the memory requirements, thus rendering the algorithm unsuitable for very large datasets.
Both of these differences were already noted in \cite{Hsieh2015}, in which the authors were able to reproduce the linear scaling behavior of AsySCD but demonstrated that it is slower than even a single threaded implementation of Algorithm \ref{alg:rcd_sync}. 

For comparison with our GPU-based implementation we have implemented the algorithm proposed in \cite{Tran2015} that uses atomic addition (A-SCD) and the ``wild'' implementation proposed in \cite{Hsieh2015} (PASSCoDE-Wild). Both implementations were written in C++ using the OpenMP library. All CPU-based experiments were run on 8-core Intel Xeon\footnote{Intel and Intel Xeon are trademarks or registered trademarks of Intel Corporation or its subsidiaries in the United States and other countries. Other product or service names may be trademarks or service marks of IBM or other companies.} CPUs with a clock frequency of 2.40GHz. Each core can run up to 2 threads resulting in a maximum number of 16 threads. 

\subsection{Twice Parallel, Asynchronous GPU Implementation}

\begin{figure}[!t]
\centering
\subfloat[Epochs]{\includegraphics[width=0.5\columnwidth]{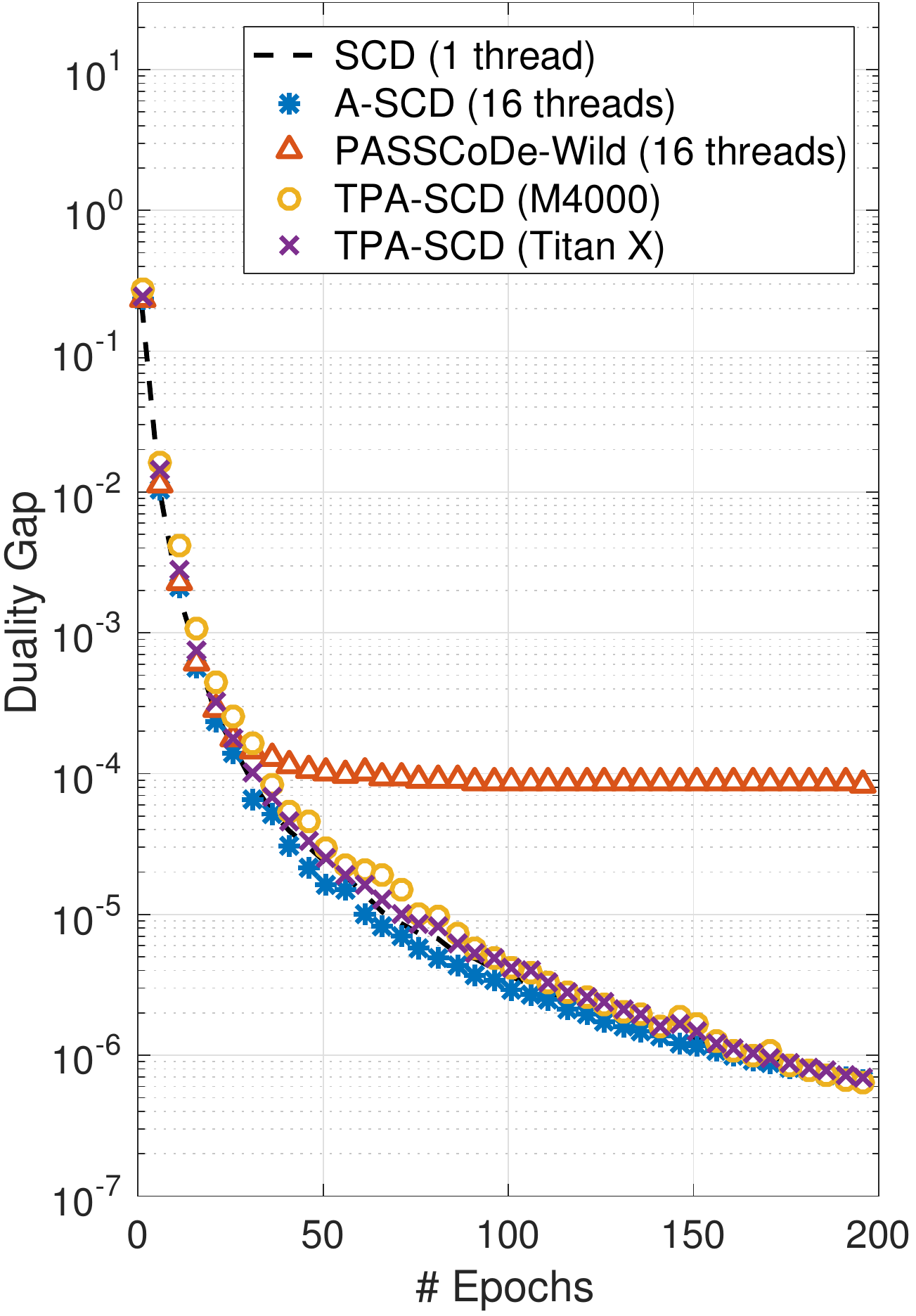}%
\label{fig:primal_epochs}}
\hfil
\subfloat[Time]{\includegraphics[width=0.5\columnwidth]{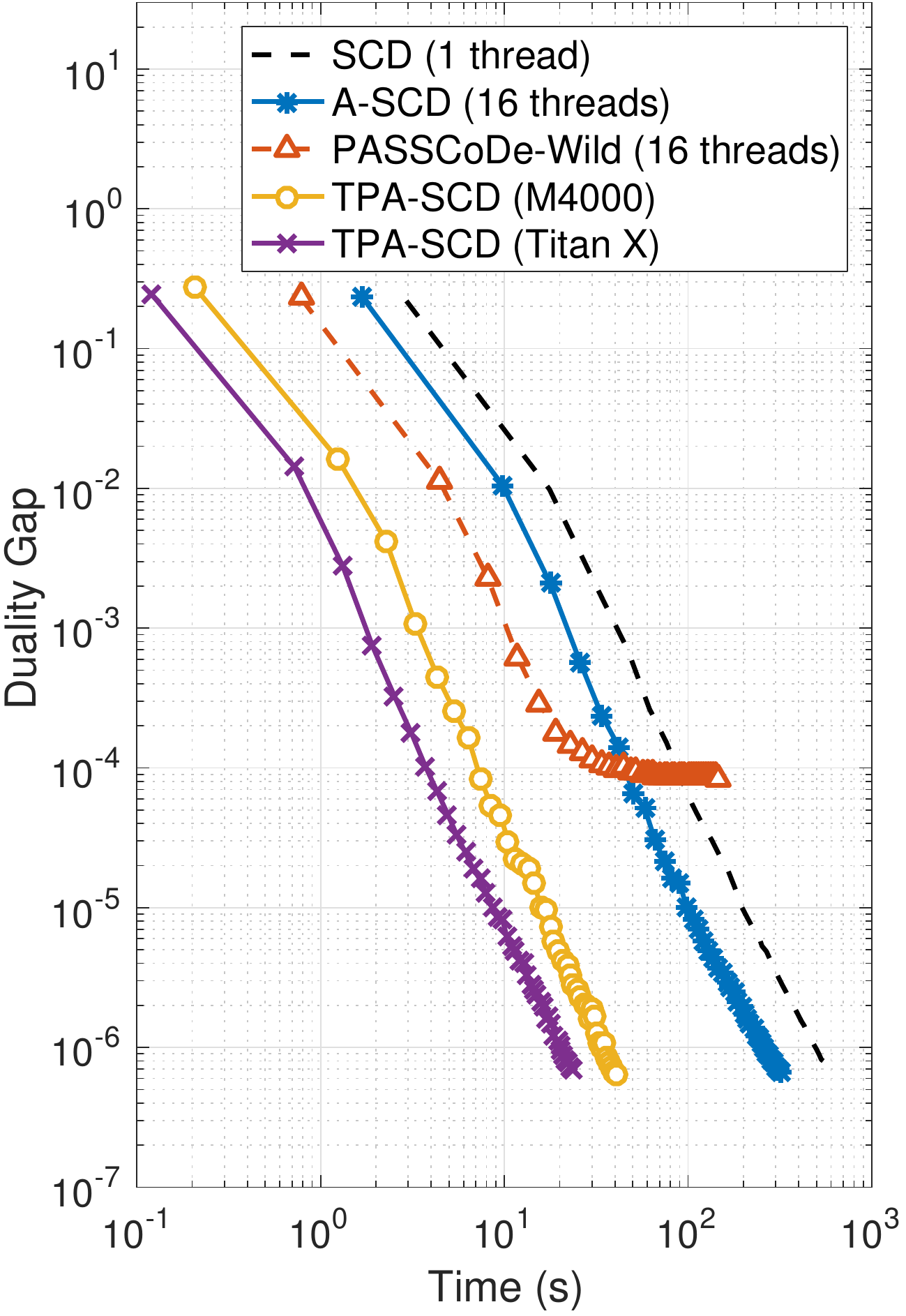}%
\label{fig:primal_time}}
\caption{Convergence in duality gap for different implementations of SCD as a function of epochs and as a function of time for the primal form of ridge regression. The webspam dataset was used with $\lambda=0.001$. }
\label{fig:primal}
\end{figure}

In Algorithm \ref{alg:scd_gpu} we present a twice parallel, asynchronous implementation of SCD  (TPA-SCD) designed to run on massively parallel GPU architectures. The algorithm is presented for the primal form of ridge regression, however the equivalent variation for the dual formulation is almost identical up to the update rule (\ref{eq:update_dual}). Our algorithm exploits two levels or parallelism. Firstly, in any given epoch, every coordinate is updated by a dedicated thread block and the thread blocks are scheduled for execution in parallel (or concurrently) on the streaming multiprocessors (SMs) that are available on the GPU. Secondly, within each thread block the computations that are required to perform the coordinate update are divided up amongst multiple threads at a very fine granularity. Furthermore, the updates to the shared vector are written out to the GPU main memory using all available threads. This helps to ensure that the shared vector and the model weights remain consistent throughout operation and thus a good convergence behavior is achieved. Rather than implement a complex locking scheme, we implemented the shared vector updates using the floating point atomic additions operations that are offered by most modern GPUs. These operations ensure that all updates to the shared vector are applied without any blocking occurring. The implementation of TPA-SCD was written in CUDA/C++ and all data is represented using $32$-bit floating point data types. We have implemented and tested the algorithm on the NVIDIA Quadro M4000 GPU as well as the GeForce GTX Titan X.

\begin{algorithm}
\caption{TPA-SCD (for GPU hardware).}
\label{alg:scd_gpu}
\begin{algorithmic}
\STATE{Initialize: $\beta=0,w=0$.}
\STATE{Copy initial vectors $\beta$ and $w$ onto GPU.}
\FOR{$epoch = 1,\ldots,n_{epochs}$}
\STATE{Generate random permutation of features $P_{epoch}$.}
\FOR{$j=1\ldots M$ (as async. GPU thread blocks)}
\FOR{$u=1\ldots n_{threads}$ (executed as warps on SM)}
%\STATE{Get shuffled coordinate:}
\IF{$u=0$}
\STATE{$m = P_{epoch}(j)$} \COMMENT{Get shuffled coordinate}
\ENDIF
%\STATE{Evaluate partial inner product:}
\STATE{$dp_u=0$} \COMMENT{Evaluate partial inner product}
\STATE{$i=u$}
\WHILE{$i < N$}
\STATE{$dp_u = dp_u + (y_i - w_i)A_{i,m}$}
\STATE{$i = i + n_{threads}$}
\ENDWHILE
\STATE{$\text{cache}[u]=dp_u$}\COMMENT{Cache in shared memory}
\STATE{synchronizeThreads()}
%\STATE{Reduce partial inner products:}
\STATE{$v=n_{threads}/2$} \COMMENT{Reduce inner products}
\WHILE{$v \neq 0$}
\IF{$u < v$}
\STATE{$\text{cache}[u] = \text{cache}[u+v]$}
\ENDIF
\STATE{synchronizeThreads()}
\STATE{$v = v/2$}
\ENDWHILE
%\STATE{Compute coordinate update:}
\IF{$u=0$} 
\STATE{$\Delta\beta_m = \left(||a_m||^2 + N\lambda\right)^{-1}\left(\text{cache}[0] - N\lambda\beta_m\right)$}
\ENDIF
\STATE{synchronizeThreads()}
%\STATE{Update shared vector:}
\STATE{$i=u$} \COMMENT{Write out updates to shared vector}
\WHILE{$i < N$}
\STATE{$w_i = w_i + A_{i,m}\Delta\beta_m$}\COMMENT{Atomic addition}
\STATE{$i = i + n_{threads}$}
\ENDWHILE
\ENDFOR
\ENDFOR
\ENDFOR
\STATE{Copy model weights $\beta$ back from GPU.}
\end{algorithmic}
\end{algorithm}

\begin{figure}[!t]
\centering
\subfloat[Epochs]{\includegraphics[width=0.5\columnwidth]{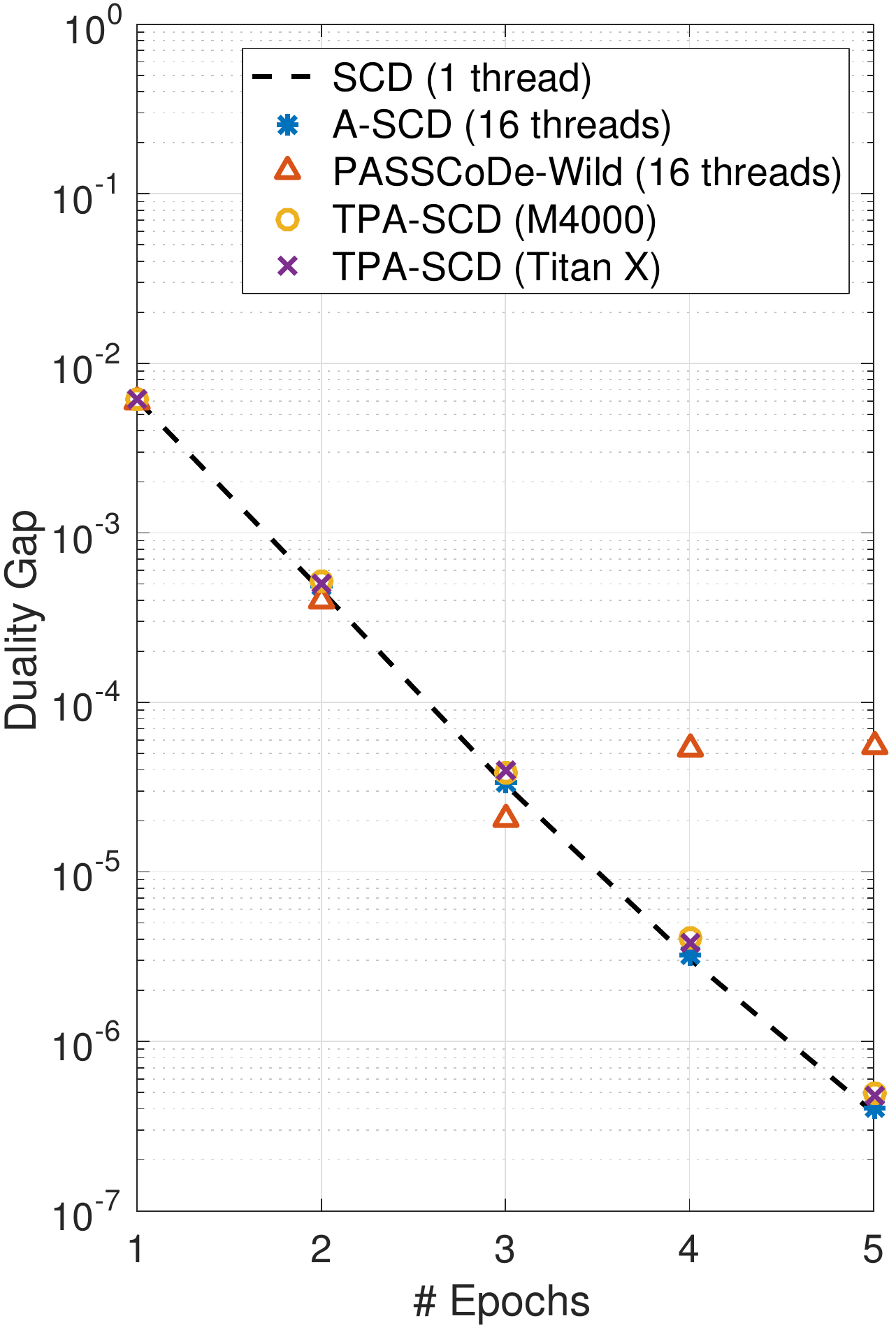}%
\label{fig:dual_epochs}}
\hfil
\subfloat[Time]{\includegraphics[width=0.5\columnwidth]{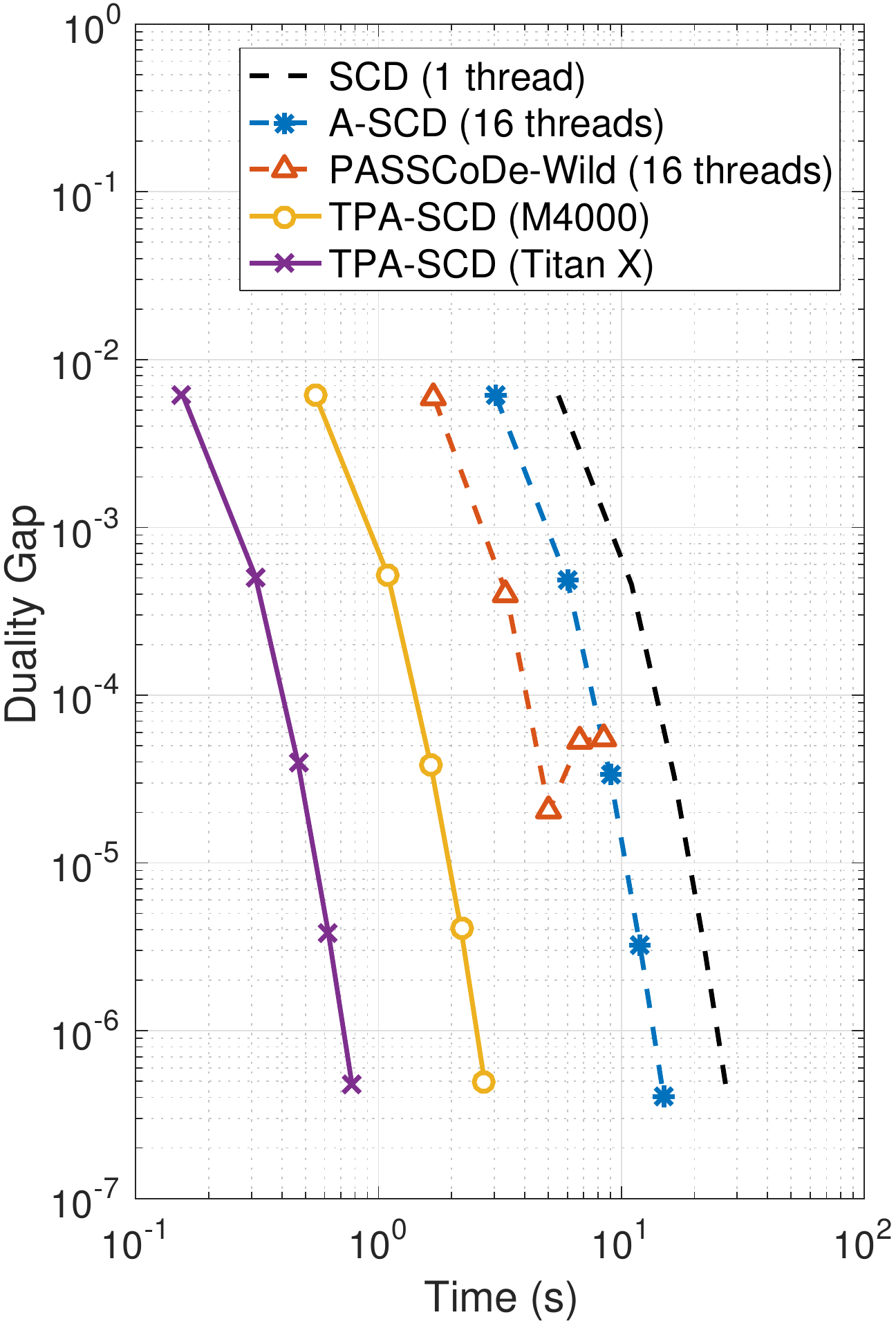}%
\label{fig:dual_time}}
\caption{Convergence in duality gap for different implementations of SCD as a function of epochs and as a function of time for the dual form of ridge regression. The webspam dataset was used with $\lambda=0.001$. }
\label{fig:dual}
\end{figure}

\subsection{Performance Comparison}

In Fig.~\ref{fig:primal} we compare the convergence behavior of a number of the algorithms that have been proposed to solve the primal form of ridge regression. The dataset that was used was a training sample of the webspam dataset \cite{Wang2012} that consists of $262,938$ examples and $680,715$ non-zero features. This sample was obtained by sampling the training examples uniformly at random to create a $75\%/25\%$ train/test split of the full dataset. A compressed sparse column format was used to represent the data matrix in the memory of the GPU when solving the primal problem and a compressed sparse row format was used when solving the dual. The sampled webspam dataset consumes around $7.3$GB of GPU memory and thus fits inside the memory capacity of the M4000 (the limit is $8$GB). In Fig.~\ref{fig:primal_epochs} we study the convergence as function of epochs and in Fig.~\ref{fig:primal_time} we compare the convergence as a function of time. When we refer to an algorithm exhibiting a ``speed up" in training time we mean that the same level of duality gap can be achieved in a shorter amount of time (even if more epochs are required). Firstly, let us consider the sequential SCD (Algorithm \ref{alg:rcd_sync}) using a single CPU thread and compare its performance with that of A-SCD and PASSCoDe-Wild (both using 16 threads). While the atomic implementation (A-SCD) has exactly the same convergence properties as the sequential algorithm as a function of epochs, we observe only a modest speed-up (around $2\times$) which we attribute to the lack of hardware support for floating point atomic addition on this particular CPU. On the other hand, for the wild implementation (PASSCoDe-Wild) we see a much more significant speed-up ($4\times$), but the algorithm converges to a solution that violates the optimality conditions (\ref{eq:duality-1}) and (\ref{eq:duality-2}). Accordingly, the duality gap does not tend towards zero. Now turning our attention to the GPU-based implementations of TPA-SCD, we observe near-perfect convergence for the algorithm on both GPUs as a function of epochs and significant gains in training time: around $14\times$ for the M4000 and $25\times$ for the Titan X. All speed-ups are measured relative to the sequential single-threaded implementation.

In Fig.~\ref{fig:dual} we compare the convergence behavior of the same set of algorithms for the dual form of ridge regression using the same dataset. From Fig.~\ref{fig:dual_epochs} we can see that things look very similar to the primal case: all implementations converge in the same manner as the sequential algorithm except PASSCoDE-Wild. The convergence behavior as a function of time is shown in Fig.~\ref{fig:dual_time}. We observe similar speed-ups for the multi-threaded CPU implementations (relative to the sequential algorithm) as were observed for the primal case. For the TPA-SCD on the M4000 we achieve a $10\times$ speed-up and on the Titan X we achieve a $35\times$ speed-up relative to the sequential SCD algorithm.

%------------------------------------------------------------------
\section{Distributed Stochastic Learning}
\label{sec:dist}
%------------------------------------------------------------------

Modern GPUs have a memory capacity of up to $16$GB thus severely limiting the size of the datasets on which we are able to learn. If we want to train on very large datasets and still benefit from the large acceleration that has been demonstrated in the previous section, it is essential that we are able to scale out the training across multiple GPUs. In this section we will review distributed SCD-based methods and introduce a technique for optimizing the aggregation of workers' model updates to accelerate convergence and improve scaling.

\subsection{Distributed SCD}

\begin{algorithm}
\caption{Distributed SCD \cite{Jaggi2014}.}
\label{alg:dist_scd}
\begin{algorithmic}
\STATE{Initialize: $w^{(0)}=0$.}
\STATE{Partition data by feature and distribute on the $K$ workers.}
\STATE{On all workers, initialize the model weights corresponding to the local features: $\beta^{(0,k)}=0$ for $k=1\ldots,K$}
\FOR{$t = 1,\ldots,n_{epochs}$}
\STATE{Broadcast $w^{(t-1)}$ to the $K$ workers.}
\FOR{$k = 1,\ldots,K$ (in parallel on workers)}
\STATE{Run one epoch of SCD on the local set of features to obtain $\beta^{(t,k)}$ and $w^{(t,k)}$.}
\STATE{Compute updates to local model weights and local version of the shared vector:}
\STATE{$\Delta\beta^{(t,k)} = \beta^{(t,k)}-\beta^{(t-1)}$}
\STATE{$\Delta w^{(t,k)} = w^{(t,k)}-w^{(t-1)}$}
\STATE{Update local model weights to ensure consistency with aggregated shared vector:}
\STATE{$\beta^{(t,k)}=\beta^{(t-1,k)} + \frac{1}{K}\Delta\beta^{(t,k)}$}
\STATE{Send $\Delta w^{(t,k)}$ to the master over the network interface.}
\ENDFOR
\STATE{Aggregate updates to shared vector on master:}
\STATE{$w^{(t)} = w^{(t-1)}+\frac{1}{K}\sum_{k}\Delta w^{(t,k)}$}
\ENDFOR
\end{algorithmic}
\end{algorithm}

Training algorithms that can be distributed across multiple machines have been the subject of a significant amount of research. Distributed techniques based on stochastic gradient descent have been proposed (see \cite{Dean2012} and \cite{Chen2016})  as well as methods based on coordinate descent/ascent (see \cite{Jaggi2014}, \cite{Takavc2015}, \cite{Mahajan2014}, \cite{Trofimov2015} and \cite{Rendle2016}). These distributed learning algorithms typically involve each machine (or worker) performing a number of optimization steps to approximately minimize the global objective function using the local data that it has available. The training data can either be distributed by sample (rows of the matrix $A$) or by feature (columns of the matrix $A$). The model updates from all of the workers are then communicated over the network to a master node. The master then aggregates all of the updates and computes a new set of model parameters. The updated model parameters are then broadcast back to the workers over the network and the process repeats. 
 
Training of ridge regression models using stochastic coordinate methods can be distributed across a cluster of machines (or a cluster of GPUs) following the aforementioned approach. One can choose whether to distribute the data matrix A across the workers by features and solve the primal form of the problem or distribute the data by example and solve the problem in its dual form. During each epoch, each worker performs a permuted pass through its set of local coordinates and performs incremental optimization of the objective function (keeping all unselected coordinates fixed, including those that exist on the other workers). The coordinate updates on each worker can be computed using any of the techniques discussed in the previous section. After all workers have finished passing through their coordinates, an aggregation step is performed whereby the updates to the shared vector on each worker are sent over the network to a master node where they are aggregated. An updated value for the shared vector is then computed on the master and broadcast back to the workers and the next epoch can begin. 

The algorithm that has been implemented in described in detail in Algorithm \ref{alg:dist_scd} for the primal formulation of ridge regression where data is distributed by features. It should be appreciated that the same procedure can be applied to dual formulation without significant modification. The procedure that is described can be thought of as a special case of the more general CoCoA framework \cite{Jaggi2014} applied specifically to the ridge regression problem (with the CoCoA hyper-parameter $\sigma$ set to $1$). The distributed aspects of the algorithm were implemented in C++ using MPI. In particular, the implementation leverages the Broadcast and Reduce functions that are offered by the Open MPI library. 

\begin{figure}[!t]
\centering
\subfloat[Primal Form]{\includegraphics[width=0.5\columnwidth]{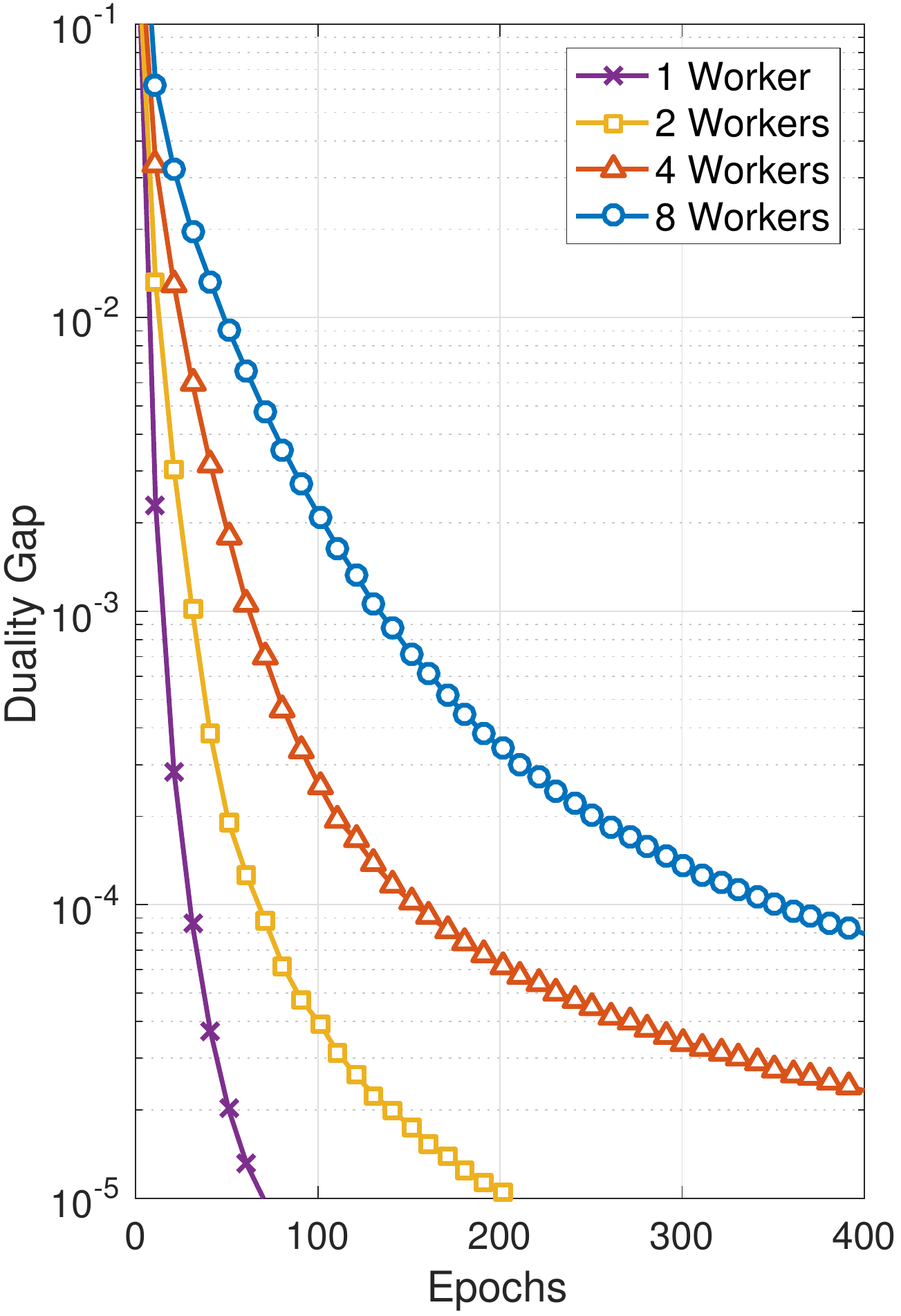}%
\label{fig:dist_gap_primal}}
\hfil
\subfloat[Dual Form]{\includegraphics[width=0.5\columnwidth]{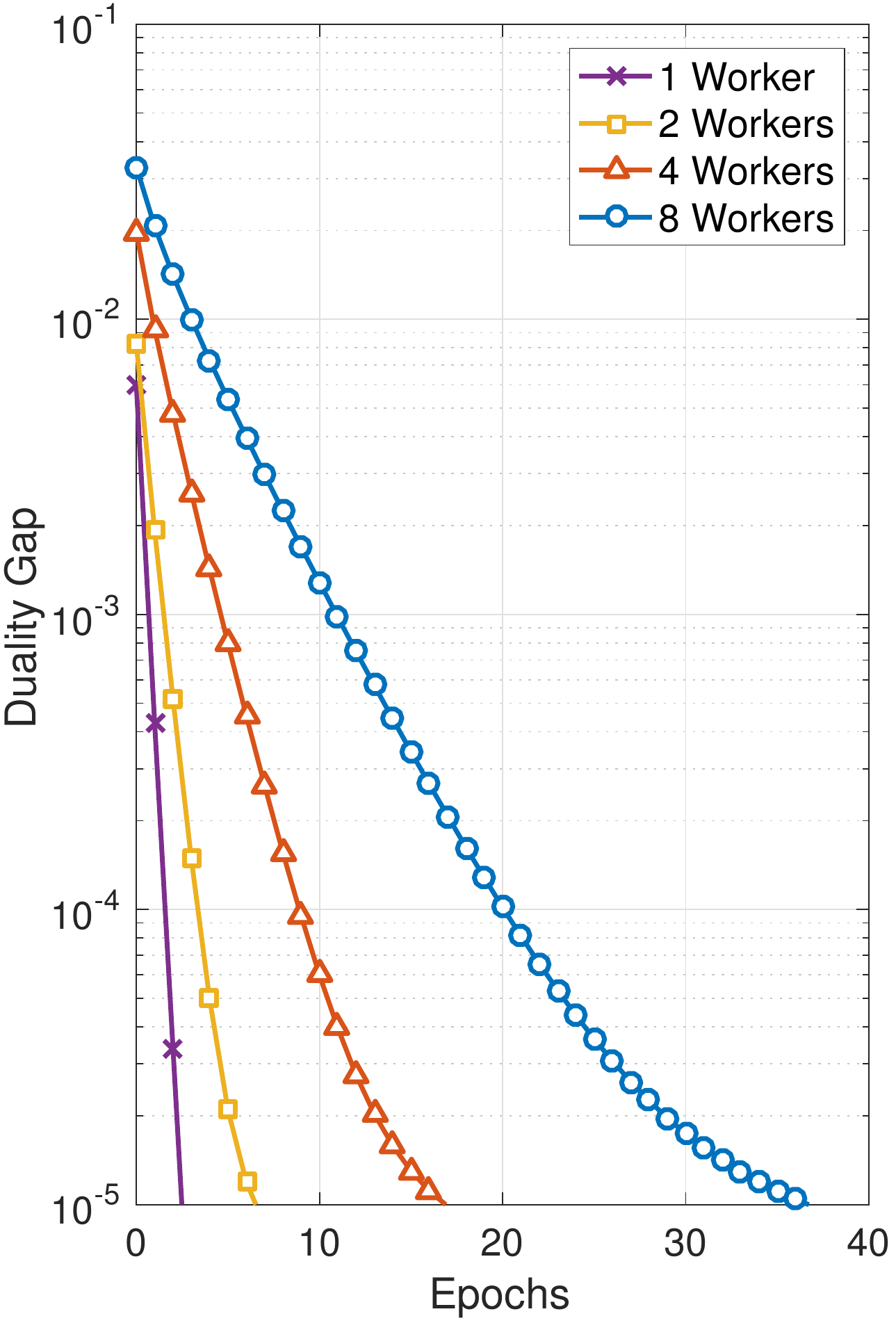}%
\label{fig:dist_gap_dual}}
\caption{Convergence in duality gap of distributed SCD for the webspam dataset with $\lambda=0.001$.}
\label{fig:dist_gap}
\end{figure}

In Fig.~\ref{fig:dist_gap} we plot the convergence in duality gap as a function of epochs for an increasing number of workers for both the primal form (where the data is distributed by features) and the dual form (where the data is distributed by examples). The experiment was run using a cluster of 4 Intel Xeon-based machines connected via a 10Gbit ethernet link with up to two workers per machine. Each worker uses the single-threaded, sequential Algorithm \ref{alg:rcd_sync} as its local solver. One can see that in both cases, the distributed algorithm converges to the optimum but there appears to be an approximately linear slow-down in convergence speed as a function of epochs. This is an inevitable effect that arises due to the workers using an out-of-date shared vector during each epoch. This effect can be somewhat alleviated if one was able to communicate shared vector updates more frequently and thus perform fewer coordinate updates on the workers between communication stages. It has been shown in \cite{Duenner2016} that there exists an infrastructure-dependent trade-off between computation and communication for distributed learning algorithms. By carefully tuning the ratio of communication to computation, it may be possible to improve the convergence behavior of the distributed algorithm further but we consider such optimizations beyond the scope of this paper.

\subsection{Adaptive Aggregation}

The convergence behavior of the distributed SCD algorithm can be improved by optimizing the aggregation step. Existing work has considered both averaging and adding of updates \cite{Ma2015}, introducing an aggregation parameter that can be set freely \cite{Smith2015} and even performing a line search method to explicitly optimize the aggregation parameter \cite{Trofimov2015}. We propose a new method to optimize aggregation for distributed ridge regression whereby an optimal value of an aggregation parameter is precisely computed in a distributed manner.

\begin{algorithm}
\caption{Distributed SCD with Adaptive Aggregation.}
\label{alg:dist_scd_adapt}
\begin{algorithmic}
\STATE{Initialize: $w^{(0)}=0,\gamma_0=1$.}
\STATE{Partition data by feature and distribute on the $K$ workers.}
\STATE{On all workers, initialize the model weights: $\beta^{(0,k)}=0$.}
\STATE{Broadcast $w^{(0)}$ to all $K$ workers.}
\FOR{$t = 1,\ldots,n_{epochs}$}
\FOR{$k = 1,\ldots,K$ (in parallel on workers)}
\STATE{Run one epoch of randomized coordinate descent on the local set of features to obtain $\beta^{(t,k)}$ and $w^{(t,k)}$.}
\STATE{Compute changes to local model and shared vector:}
\STATE{$\Delta\beta^{(t,k)} = \beta^{(t,k)}-\beta^{(t-1)}$}
\STATE{$\Delta w^{(t,k)} = w^{(t,k)}-w^{(t-1)}$}
\STATE{Compute $||\Delta\beta^{(t,k)}||^2$ and $\left<\beta^{(t,k)},\Delta \beta^{(t,k)}\right>$.}
\ENDFOR
\STATE{Aggregate updates on master:}
\STATE{$\Delta w^{(t)}=\sum_{k}\Delta w^{(t,k)}$}
\STATE{$\left<\beta^{(t)},\Delta \beta^{(t)}\right> = \sum_{k=1}^K \left<\beta^{(t,k)},\Delta \beta^{(t,k)}\right>$}
\STATE{$||\Delta\beta^{(t)}||^2 =  \sum_{k=1}^K ||\Delta\beta^{(t,k)}||^2$}
\STATE{Compute optimal aggregation $\gamma_t$ parameter using (\ref{eq:opt_agg}).}
\STATE{Apply aggregated updates to the shared vector:}
\STATE{$w^{(t)} = w^{(t-1)}+\gamma_t\Delta w^{(t)}$}
\STATE{Broadcast $w^{(t)}$ and $\gamma_t$ to all $K$ workers.}
\FOR{$k = 1,\ldots,K$ (in parallel on workers)}
\STATE{Update local model weights for consistency:}
\STATE{$\beta^{(t,k)} = \beta^{(t-1,k)}+\gamma_t\Delta \beta^{(t,k)}$}
\ENDFOR
\ENDFOR
\end{algorithmic}
\end{algorithm}

Let us denote the aggregated model weights and shared vector at the end of epoch $t$ as follows:
\begin{eqnarray}
\beta^{(t+1)} &=& \beta^{(t)} + \gamma_t\sum_{k=1}^K\Delta \beta^{(t,k)}  =  \beta^{(t)} + \gamma_t\Delta \beta^{(t)},\nonumber\\
w^{(t+1)} &=& w^{(t)} + \gamma_t\sum_{k=1}^K\Delta w^{(t,k)} = w^{(t)} + \gamma_t\Delta w^{(t)}, \nonumber
\end{eqnarray}
where $\gamma_t$ is the aggregation parameter in the $t$-th epoch.
For the primal formulation of ridge regression, we can then optimize the objective function to explicitly find the best aggregation parameter at every epoch:
\begin{eqnarray}
\gamma_t^{(*)}=\argmin_{\gamma}\mathcal{P}\left(\beta^{(t)} + \gamma\Delta \beta^{(t)},w^{(t)} + \gamma\Delta w^{(t)} \right).\nonumber
\end{eqnarray}
The above equation has the following explicit solution:
\begin{eqnarray}
\gamma_t^{(*)}= -\frac{\left(\left<w^{(t)},\Delta w^{(t)}\right>+N\lambda\left<\beta^{(t)},\Delta \beta^{(t)}\right>\right)}{||\Delta w^{(t)}||^2+N\lambda||\Delta\beta^{(t)}||^2}.\label{eq:opt_agg}
\end{eqnarray}
While the aggregated changes to the shared vector $\Delta w^{(t)}$ are already available on the master node, the aggregated changes to the model weights $\Delta \beta^{(t)}$ are not. However, since all workers only update the coordinates corresponding to their local data, the following property allows us to compute $\left<\beta^{(t)},\Delta \beta^{(t)}\right>$ and $||\Delta\beta^{(t)}||^2$ in a distributed manner:
\begin{eqnarray}
\left<\beta^{(t)},\Delta \beta^{(t)}\right> &=& \sum_{k=1}^K \left<\beta^{(t,k)},\Delta \beta^{(t,k)}\right>, \nonumber \\
||\Delta\beta^{(t)}||^2 &=&  \sum_{k=1}^K ||\Delta\beta^{(t,k)}||^2 .\nonumber
\end{eqnarray}
The distributed algorithm is defined precisely in Algorithm \ref{alg:dist_scd_adapt} for the primal form of ridge regression. The additional communication that is introduced in order to achieve the optimized aggregation amounts to the transfer of a few scalars over the network interface per epoch. The equivalent algorithm for the dual form follows easily from the following expression for the optimal aggregation parameter in the dual setting:
\begin{eqnarray}
\bar{\gamma}_t^{(*)} = \frac{\left<\Delta\alpha^{(t)},y\right>-N\left<\Delta\alpha^{(t)},\alpha^{(t)}\right>-\frac{1}{\lambda}\left<\Delta \bar{w}^{(t)},\bar{w}^{(t)}\right>}{\frac{1}{\lambda}||\Delta \bar{w}^{(t)}||^2+N||\alpha^{(t)}||^2}.\nonumber
\end{eqnarray}

\begin{figure}[!t]
\centering
\subfloat[Primal Form]{\includegraphics[width=0.5\columnwidth]{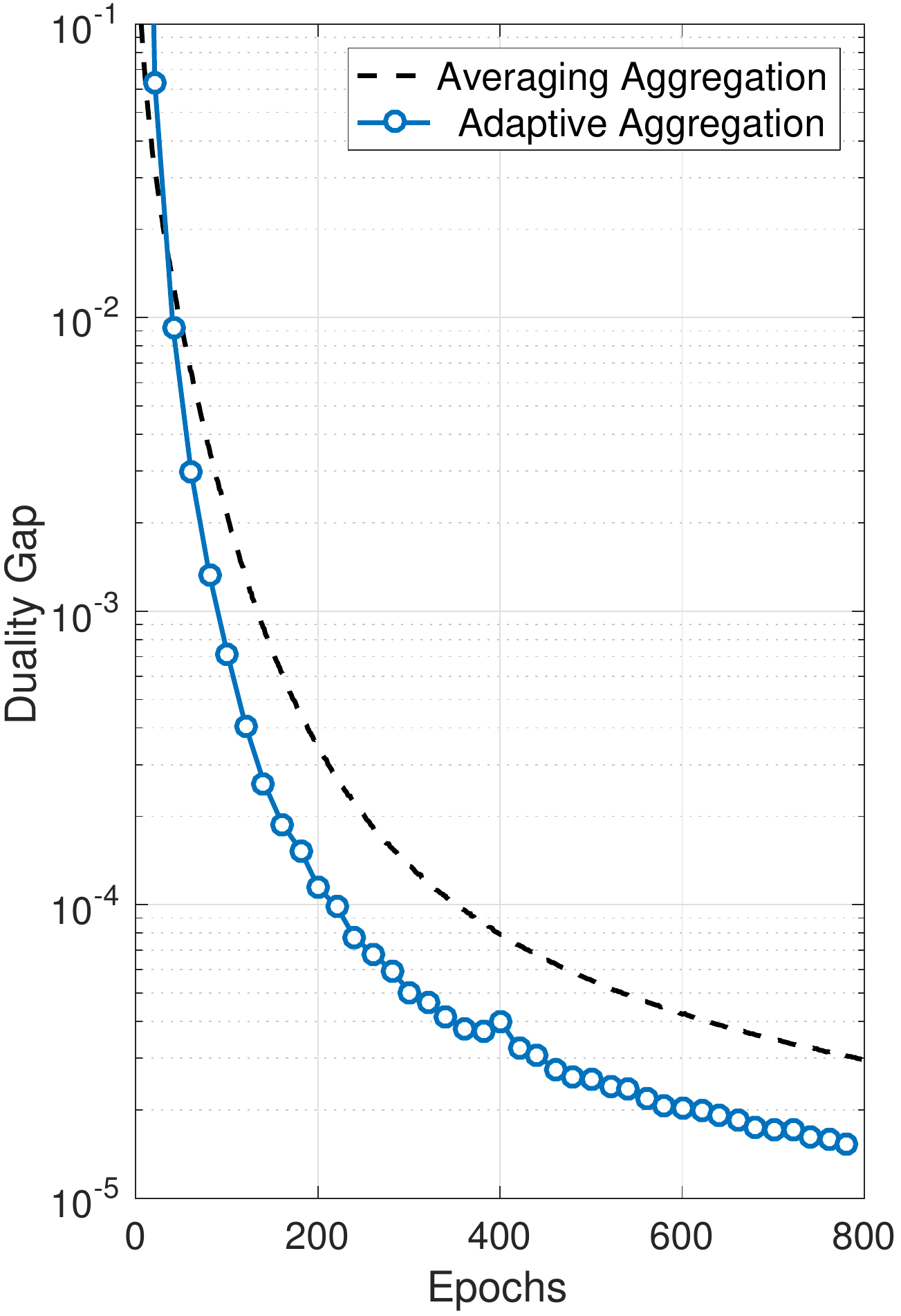}%
\label{fig:dist_adapt_primal}}
\hfil
\subfloat[Dual Form]{\includegraphics[width=0.5\columnwidth]{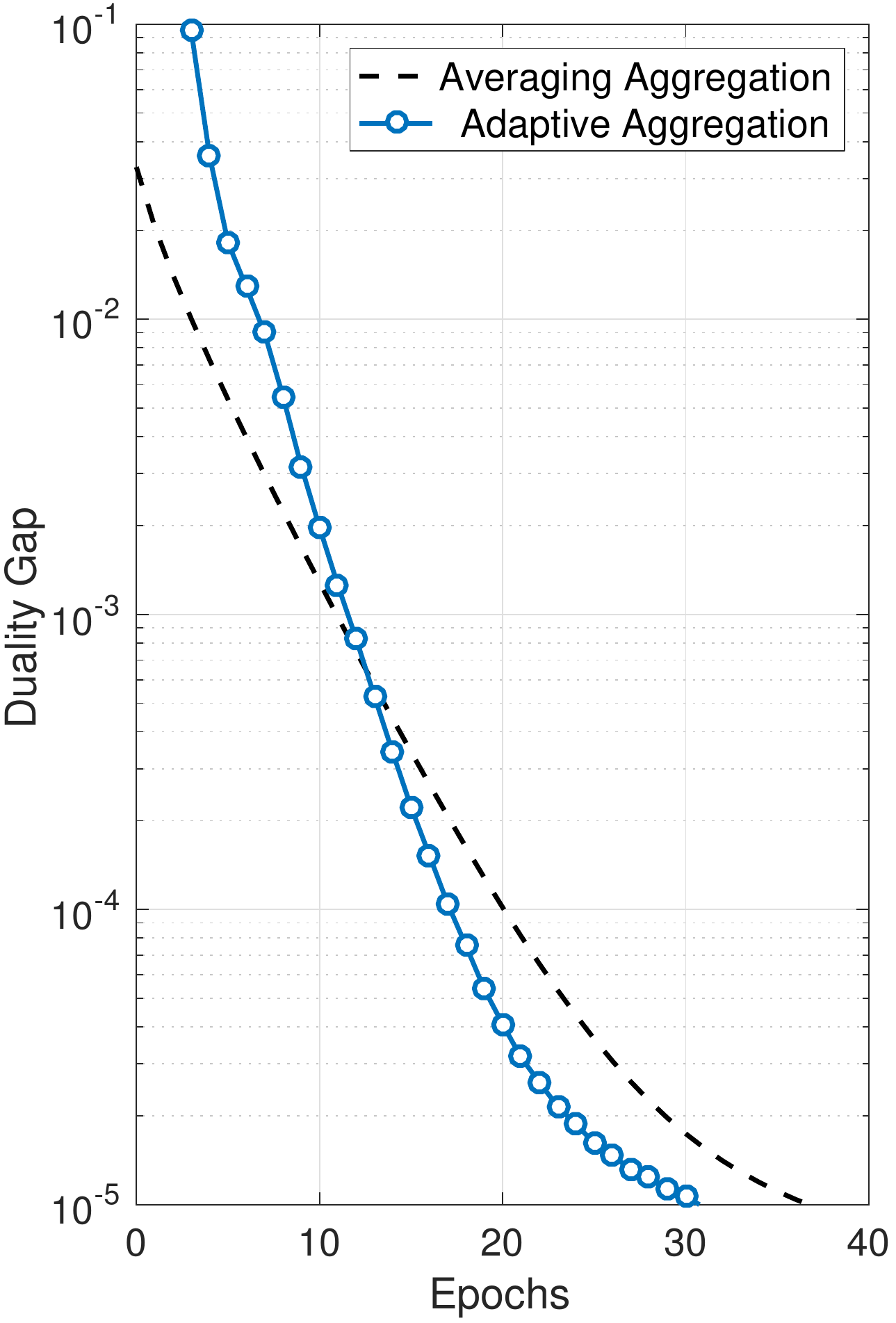}%
\label{fig:dist_adapt_dual}}
\caption{The effect of adaptive aggregation on distributed SCD for the webspam dataset with $\lambda=0.001$ with $K=8$ workers.}
\label{fig:dist_adapt}
\end{figure}

\begin{figure}[!t]
\centering
\subfloat[Primal Form]{\includegraphics[width=0.5\columnwidth]{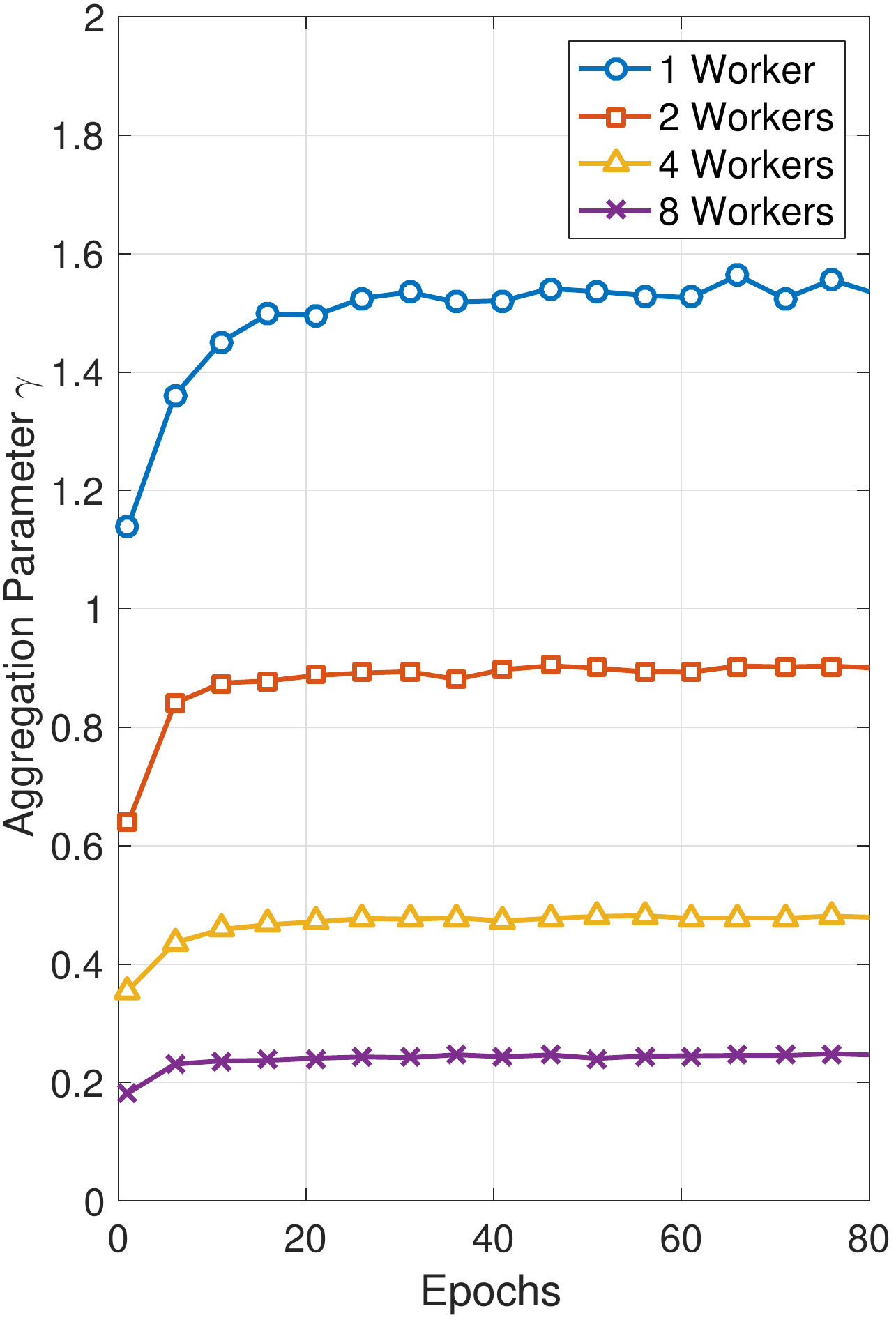}%
\label{fig:dist_gamma_primal}}
\hfil
\subfloat[Dual Form]{\includegraphics[width=0.5\columnwidth]{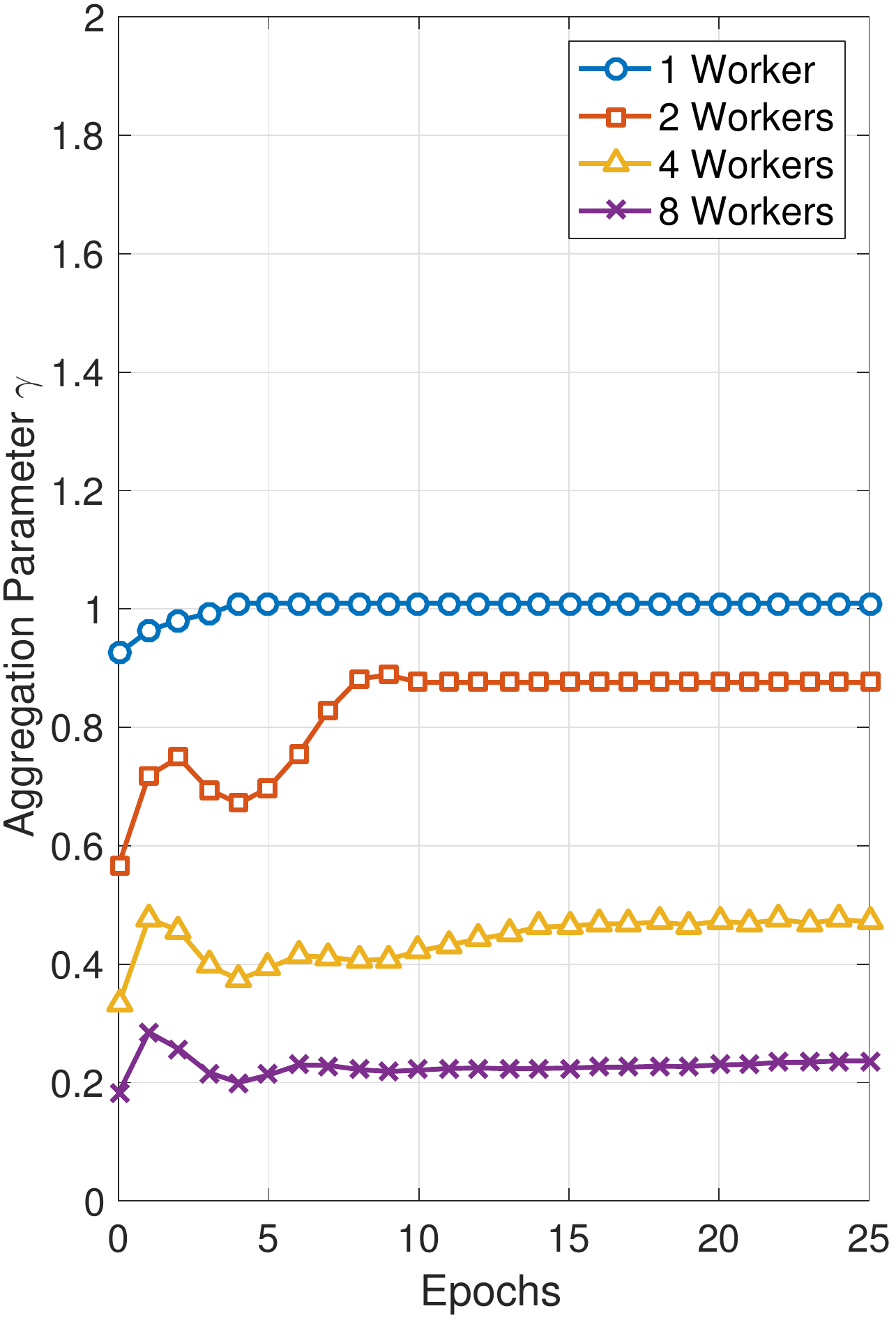}%
\label{fig:dist_gamma_dual}}
\caption{The evolution of the optimal aggregation parameter for the webspam dataset with $\lambda=0.001$.}
\label{fig:dist_gamma}
\end{figure}

In Fig.~\ref{fig:dist_adapt} we plot the convergence behavior of distributed SCD using adaptive aggregation on the ridge regression problem and compare it with the algorithm that uses averaging for aggregating the updates to the shared vector. We observe that for the algorithm that solves the primal formulation there is a speed-up in convergence that approaches $2\times$ for small values of duality gap. For the algorithm that solves the dual, the effect is less pronounced: for relatively large values of the duality gap the algorithm with adaptive aggregation can be slower (since we explicitly minimize the dual objective, the duality gap is not necessarily minimized) but for small values of the duality gap we observe an speed-up of around $1.2\times$. 

In Fig.~\ref{fig:dist_gamma} we show the evolution of the optimal value of the aggregation parameter as a function of epochs. We can observe a trend: it tends to start off relatively low before increasing and finally converging to some value. It is interesting to note that the value to which it converges to is significantly larger than the value that corresponds to averaging (i.e., $\gamma=1/K$). 

\begin{figure}[!t]
\centering
\subfloat[Primal Form]{\includegraphics[width=0.5\columnwidth]{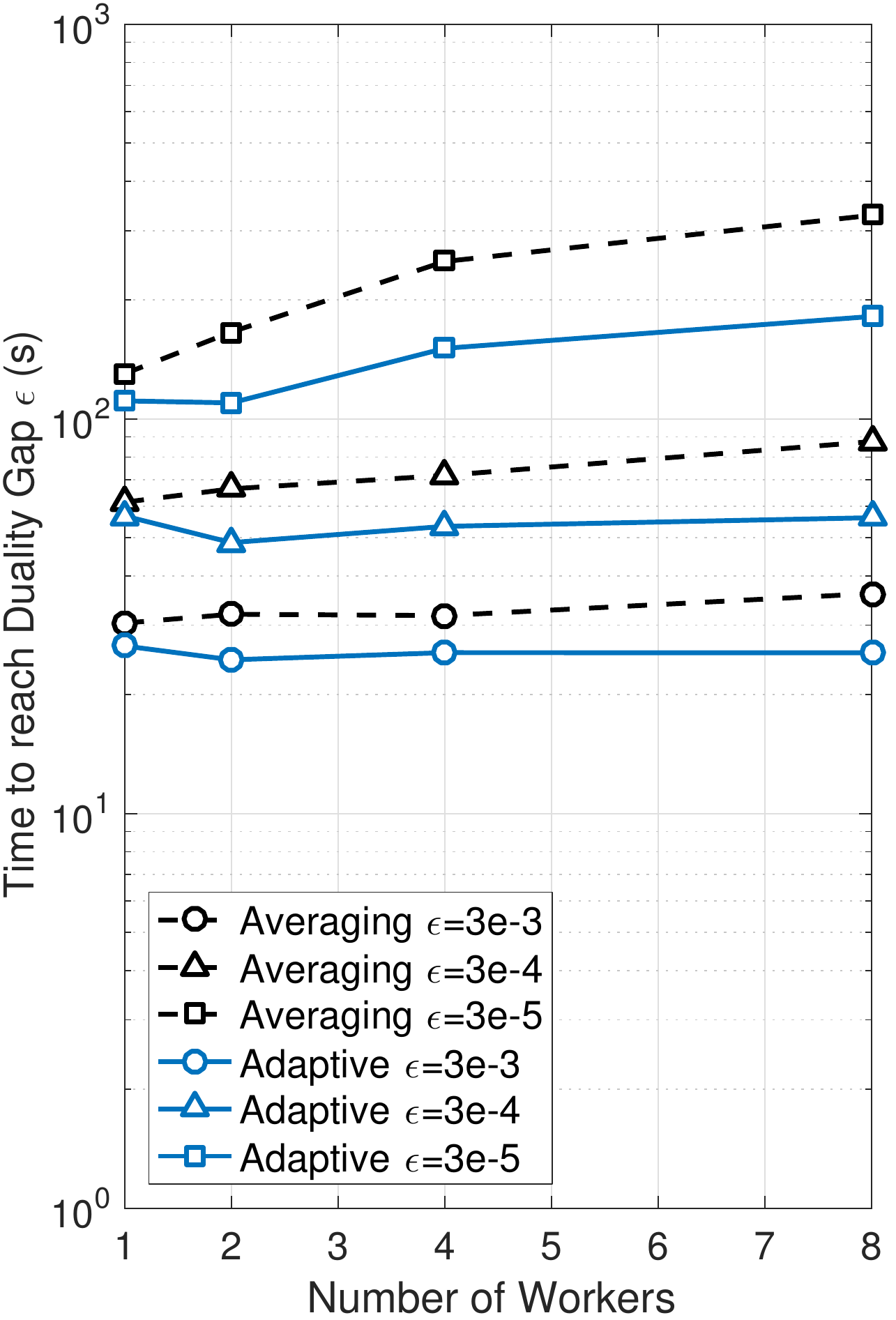}%
\label{fig:dist_scale_primal}}
\hfil
\subfloat[Dual Form]{\includegraphics[width=0.5\columnwidth]{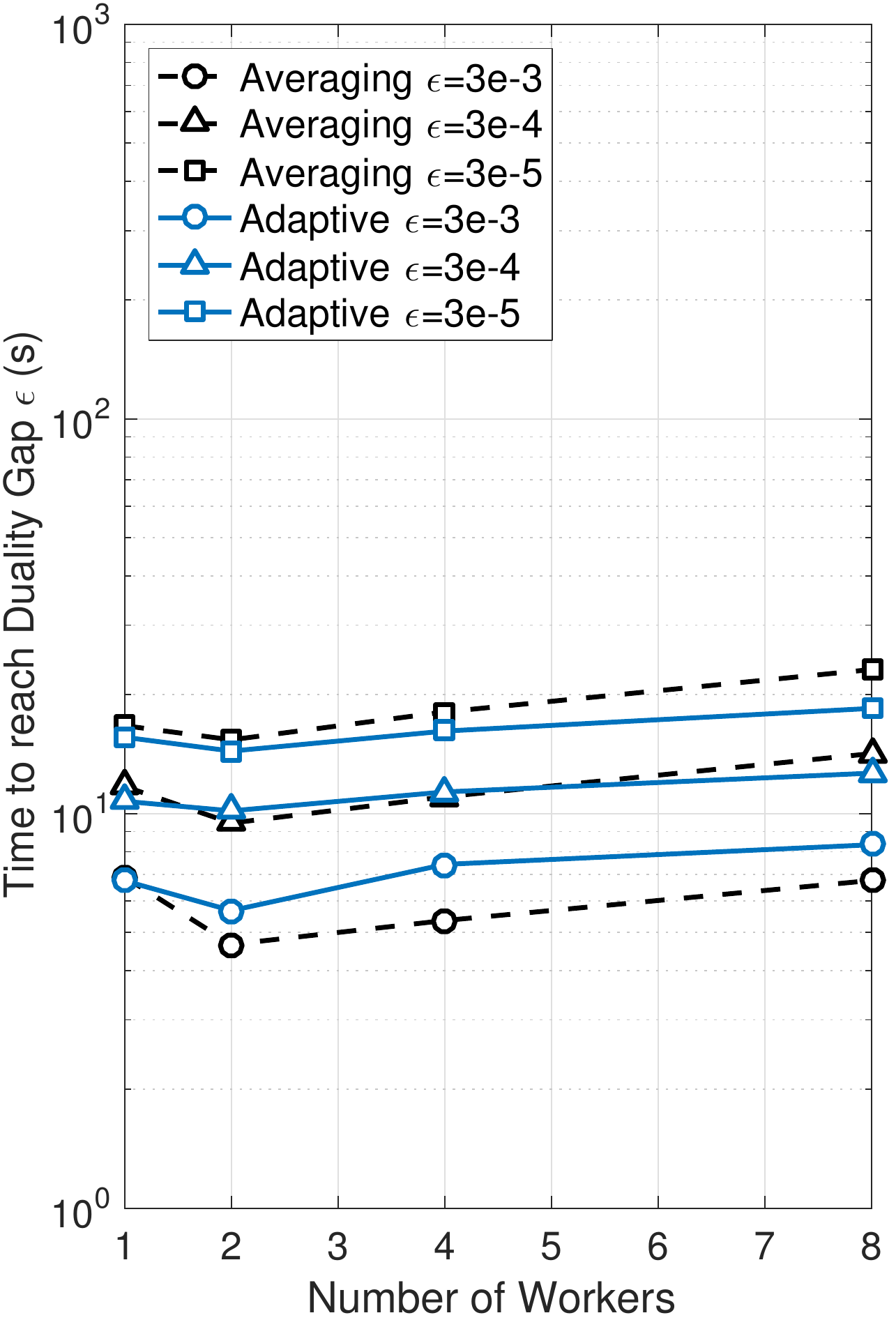}%
\label{fig:dist_scale_dual}}
\caption{Time to reach a target duality gap for distributed SCD on the webspam dataset with $\lambda=0.001$.}
\label{fig:dist_scale}
\end{figure}

In Fig.~\ref{fig:dist_scale} we plot the time to reach a desired duality gap as a function of the number of workers for the webspam dataset. In Fig.~\ref{fig:dist_scale_primal} we show the scaling behavior for the distributed solver for the primal form of ridge regression and in Fig.~\ref{fig:dist_scale_dual} we show the same but for the dual formulation. In both cases we can compare the scaling behavior, for different levels of desired accuracy, with and without the adaptive aggregation technique. In both cases we observe that the adaptive aggregation technique allows us to scale out across multiple worker nodes while keeping the training time roughly constant. For the dual problem on the webspam dataset, we see that for relatively high values of the duality gap, the adaptive aggregation can slow down convergence somewhat. This is consistent with Fig.~\ref{fig:dist_adapt_dual} where we observed a crossover point at around duality gap $5\times 10^{-4}$.

This scaling behavior is very consistent with that reported for CoCoA+ in \cite{Ma2015}. The acceleration that comes from each worker processing a factor of $K$ less data per iteration is just enough to compensate for the linear slow-down in convergence that occurs due to each worker using an outdated model (see Fig.~\ref{fig:dist_gap}). The scaling behavior strongly depends on the nature of the underlying dataset. In particular, the slow-down in convergence is determined by the level of correlation between coordinates on the different workers. If there exists some additional structure (for instance, a large number of one-hot encoded categorical variables) then one can partition the coordinates in an intelligent way to achieve a faster convergence and thus better scaling \cite{Rendle2016}.

\section{Scaling out across multiple GPUs}
\label{sec:dist_gpu}

\begin{figure}
\centering
\includegraphics[width=1.0\columnwidth]{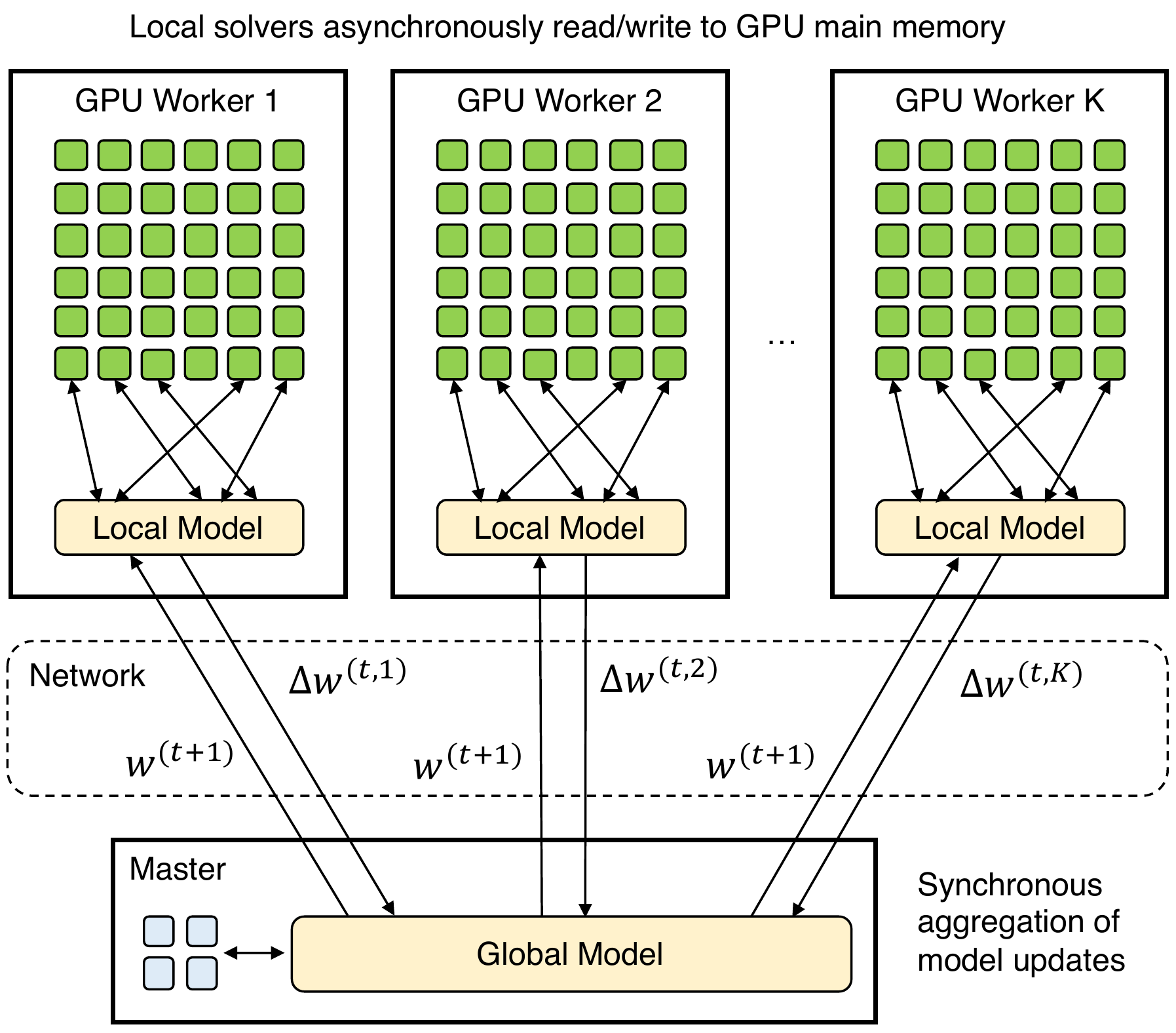}
\caption{Distributed learning across a cluster of GPUs.}
\label{fig:dist_gpu_arch}
\end{figure}

In this section we will combine the methods of Sections \ref{sec:gpu} and \ref{sec:dist} to construct an accelerated implementation of TPA-SCD that can scale across multiple GPUs connected over a network and train on datasets much larger than the memory capacity of single GPU device.

\subsection{Distributed TPA-SCD}
The general approach is illustrated in Fig.~\ref{fig:dist_gpu_arch}. The training is distributed across $K$ workers using the algorithms described in the previous section. Each worker consists of a CPU-based machine with at least one GPU attached over a PCIe interface. During each epoch, every worker runs the TPA-SCD algorithm on the streaming multiprocessors of its GPU and computes updates to its local model weights as well the shared vector. Each worker is then responsible for copying the shared vector updates from the GPU device memory into its host memory and then communicating the updates to the master over the network interface. The master then aggregates the updates and broadcasts the new shared vector back to the workers. The worker must then copy the new shared vector from its host memory back into the GPU device memory. Thus, we have opted to use synchronous communication between the workers at the network level and asynchronous communication between the ``sub-workers" at the GPU level (i.e., the thread blocks that are processing different coordinates). Note that the dataset on which we are training is transferred into the GPU memory once at the beginning of operation and does not move. Thus the penalties associated with transferring large amount of data over the network are for the most part avoided. Communication of vectors on and off the GPU during each epoch was implemented using the pinned memory functionality offered by CUDA to achieve maximum throughput over the PCIe interface between the workers' host memory and device memory.

\begin{figure}[!t]
\centering
\subfloat[NVIDIA Quadro M4000]{\includegraphics[width=0.5\columnwidth]{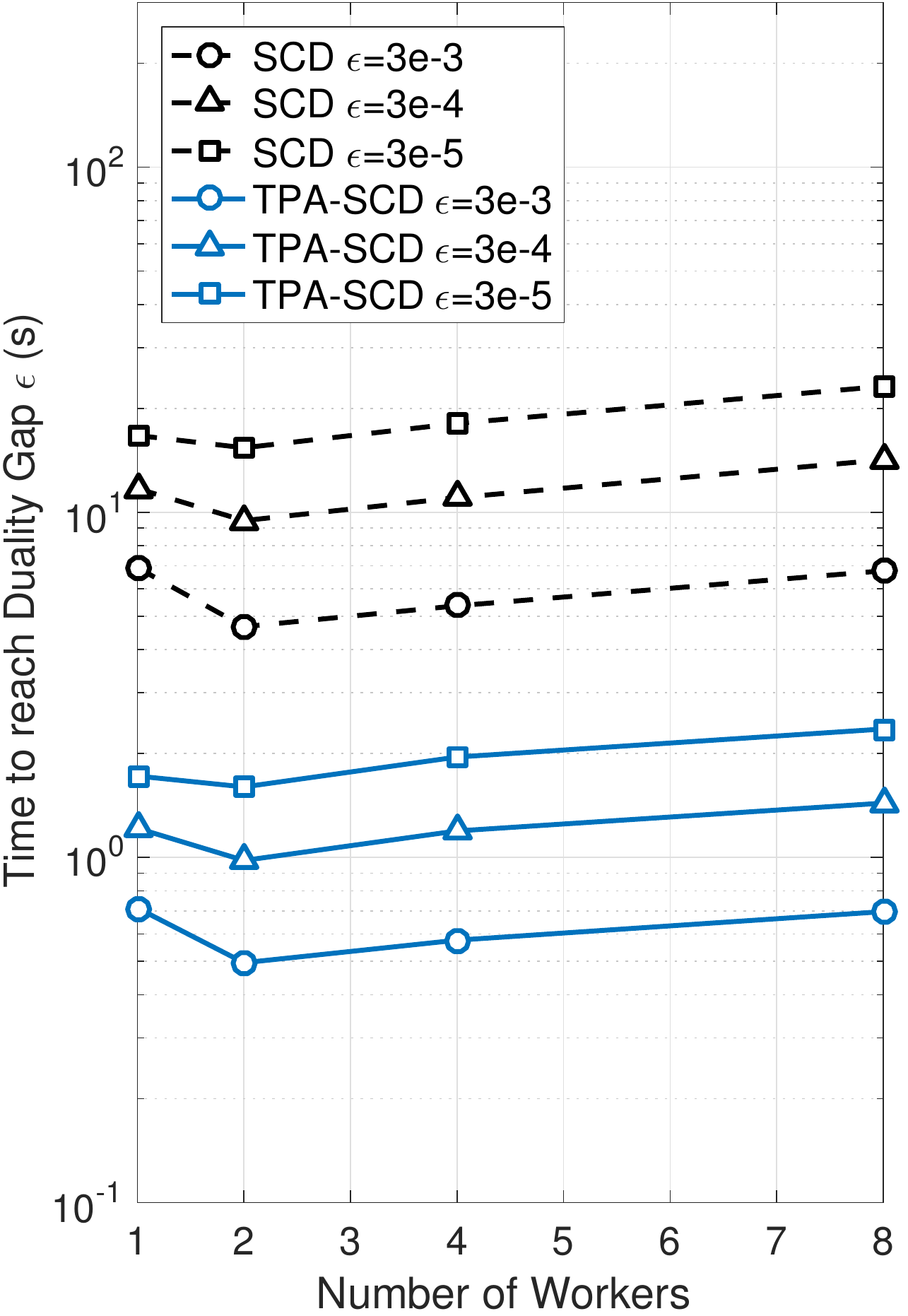}%
\label{fig:dist_gpu_m4000}}
\hfil
\subfloat[GeForce GTX Titan X]{\includegraphics[width=0.5\columnwidth]{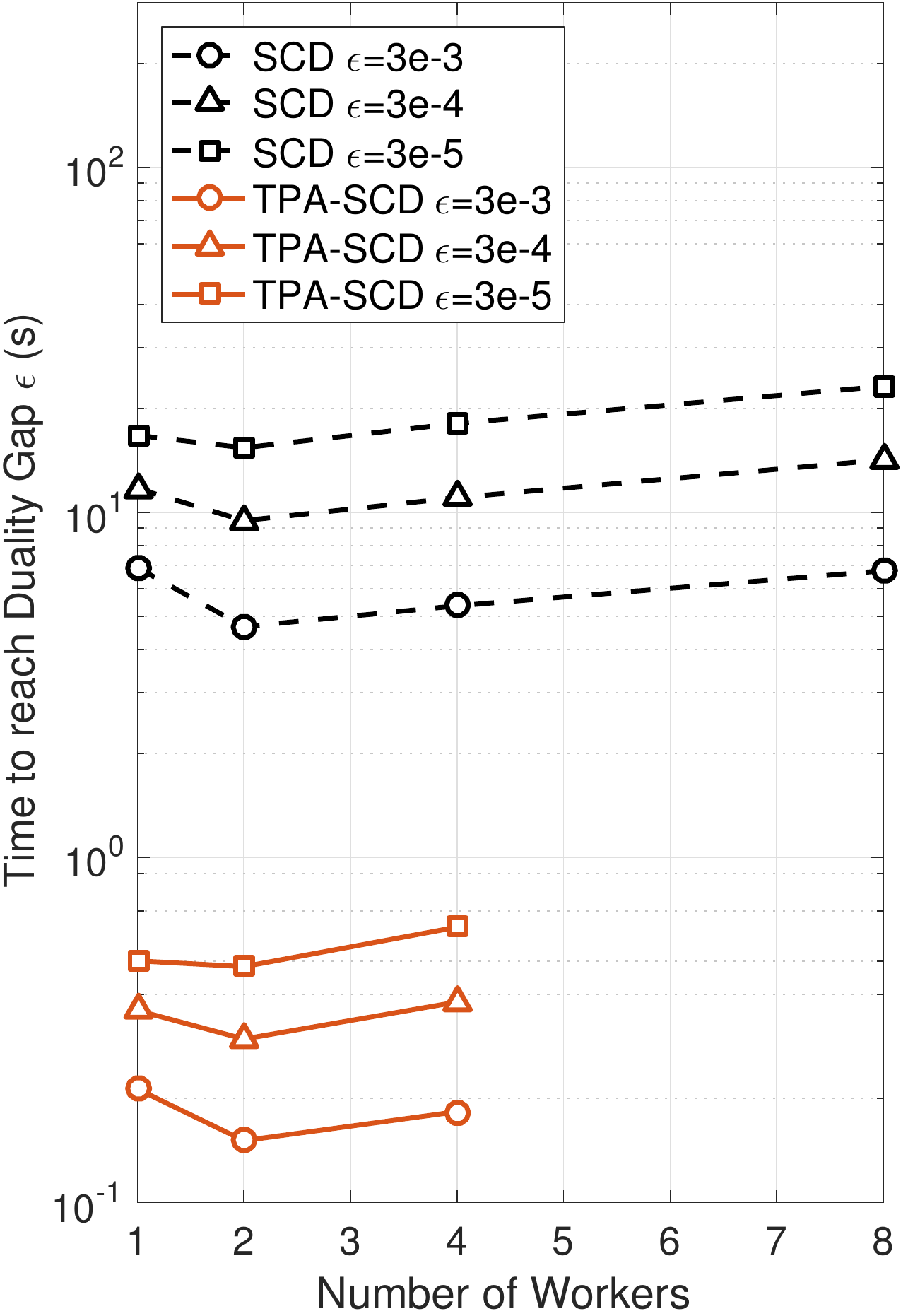}%
\label{fig:dist_gpu_titanx}}
\caption{Scaling out ridge regression in its dual form across two different clusters of GPUs. The webspam dataset was used with $\lambda=0.001$.}
\label{fig:dist_gpu}
\end{figure}

\begin{figure}[!t]
\centering
\includegraphics[width=\columnwidth]{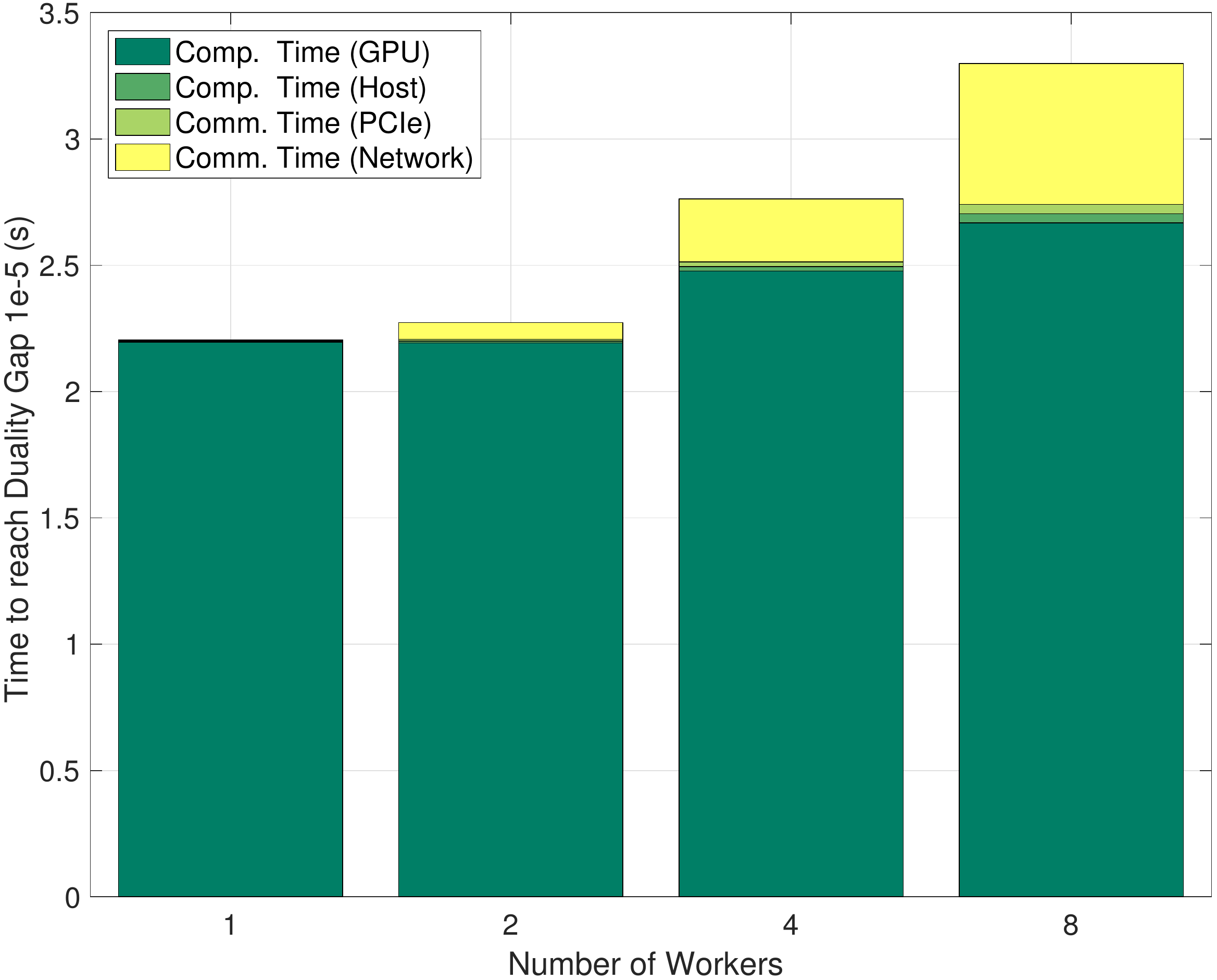}
\caption{Computation vs. communication on the M4000 cluster for ridge regression in its dual form. The webspam dataset was used with $\lambda=0.001$.}
\label{fig:dist_gpu_comm}
\end{figure}

In Fig.~\ref{fig:dist_gpu} we show the scaling behavior of distributed TPA-SCD (with averaging) for the webspam dataset using two different GPU clusters. The dual formulation of ridge regression is being solved and the data is thus distributed across the GPU memory by training examples. In Fig.~\ref{fig:dist_gpu_m4000} we have used a cluster of eight NVIDIA Quadro M4000 GPUs that are connected via a 10Gbit ethernet network link.  We observe a $10\times$ speed-up over the equivalent distributed implementation that uses sequential SCD. In Fig.~\ref{fig:dist_gpu_titanx} we show results using a cluster of 4 GeForce GTX Titan X GPUs that are attached to a single machine and communicate over the PCIe interface. These GPUs are significantly faster than the M4000s and we observe around a $30\times$ speed-up and similar scaling behavior. Note that in these results we have not applied the adaptive aggregation technique and thus all speed-ups reported are solely due to execution of the local solver on the GPU hardware. 

In Fig.~\ref{fig:dist_gpu_comm} we examine the scaling behavior on the M4000 cluster in more detail. The execution time is broken down into the time spent computing (both on the GPU and on the host), the time spent transferring data on/off the GPU over PCIe and the time spent communicating over the 10Gbit ethernet network. 
While we observe that the time spent computing on the GPU dominates the execution time in all cases, we notice that the communication overheads increase as the number of workers grows.
However, with 8 workers the communication time is still only around $17\%$ of the total execution time, suggesting that it should be possible to scale out across more workers before the communication overheads become prohibitive. Naturally, these results indicate that the use of a 100Gbit ethernet network interface would improve the scaling behavior further. We would like to stress that the scaling behavior that has been demonstrated does not imply that training can be accelerated if the size of the dataset remains fixed. However, as we will now demonstrate, this scaling property allows one to leverage GPU acceleration when training massive datasets that do not fit inside the memory of a single GPU.

\subsection{Large-scale data}

While the speed-ups we observe for the webspam dataset are consistent with the results reported in Fig.~\ref{fig:dual_time} using a single GPU, we now have the ability to train using much larger datasets that do not fit inside the memory of a single GPU. For our next experiment we sampled one day's worth of data from the criteo dataset \cite{criteo2015}. This sample consists of approximately 200 million training examples and 75 million unique features and occupies around 40GB of GPU memory using a compressed sparse row format\footnote{For this sample the values in the training data matrix are always $1$ and so one could halve the memory usage by re-writing the code to explicitly assume this. Even so, the dataset would not fit in the memory of a single GPU}. 

We partition the dataset by training example and thus randomly distribute the rows of the training data matrix across the 4 workers of the Titan X GPU cluster. We then ran distributed TPA-SCD with adaptive aggregation and compared the convergence behavior (as a function of time) with that of two reference distributed implementations. The first reference implementation is distributed SCD (Algorithm \ref{alg:dist_scd}) using 4 workers. Each worker uses single-threaded, sequential SCD as its local solver. The second reference is the same except all workers use PASSCoDe-Wild (with 16 threads) as their local solver. We have decided not to compare with the distributed implementation consisting of 64 single-threaded workers (i.e., 16 workers on each of the 4 CPUs) since it has been established in Section \ref{sec:dist} that using more workers does not lead to faster convergence.

\begin{figure}[!t]
\centering
\includegraphics[width=\columnwidth]{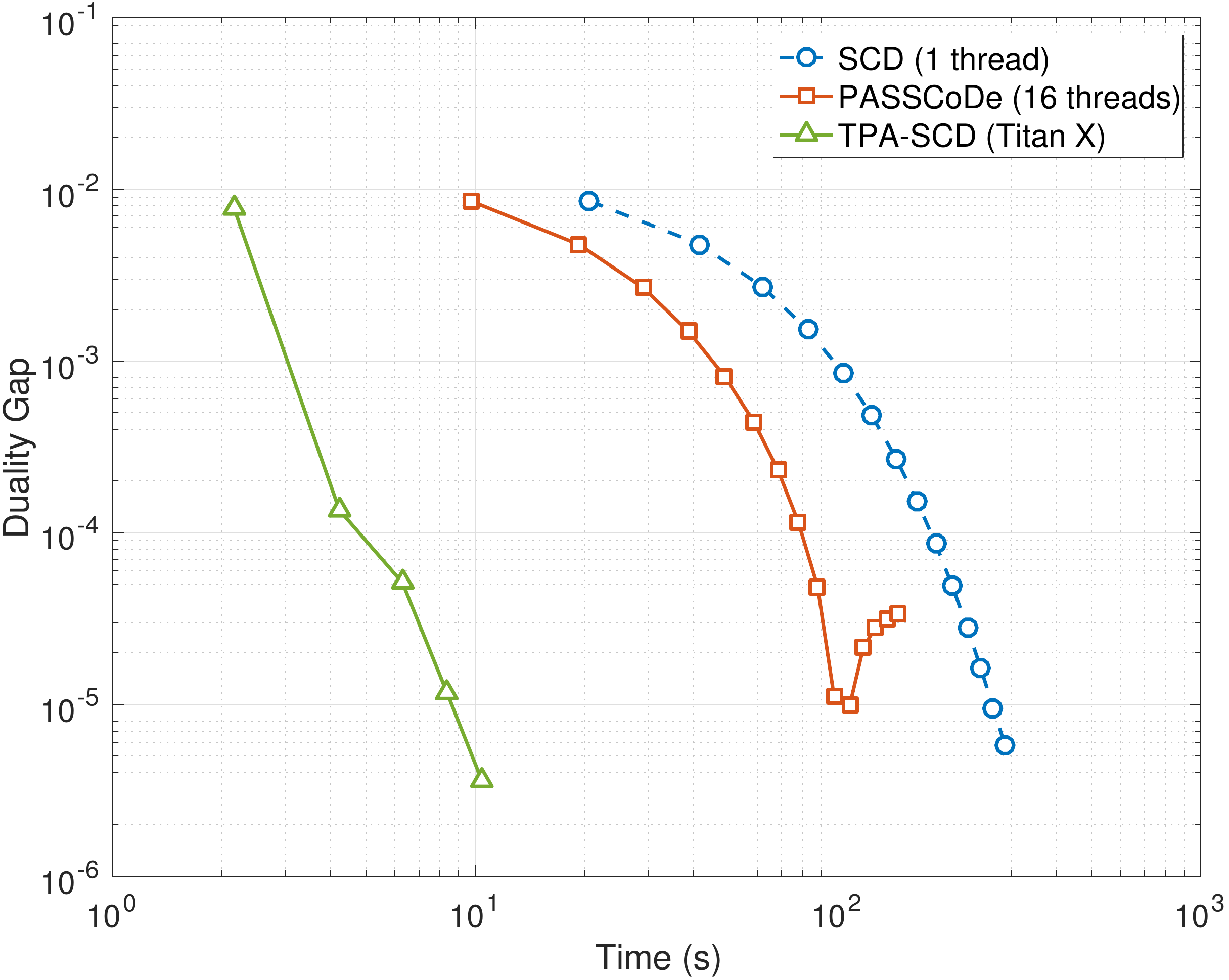}
\caption{Convergence in duality gap as a function of time for distributed SCD-based algorithms on a 40GB sample of the criteo dataset. 4 workers are used in all cases.}
\label{fig:dist_gpu_criteo}
\end{figure}

The convergence in duality gap for these three schemes is presented in Fig.~\ref{fig:dist_gpu_criteo}. We see around a $40\times$ speed-up relative to the single-threaded SCD and around a $20\times$ speed-up relative to PASSCoDe-Wild. Note that since the optimality conditions are violated by the multi-threaded CPU implementation, the duality gap does not converge to zero. However, the solution that it has found may still be useful depending on the application.

\section{Conclusion}

In this work we have presented a new implementation of stochastic coordinate descent (TPA-SCD) that has been carefully designed to efficiently make use of the compute architecture provided by modern GPUs. We have demonstrated that GPUs can be used to train a ridge regression model to a desired degree of accuracy $35\times$ faster than a single-threaded CPU implementation and $10\times$ faster than a multi-threaded CPU implementation. In order to scale up to very large datasets that consist of hundreds of millions of training examples and features we have demonstrated that it is possible to scale out our stochastic learning system across 8 GPUs without any significant loss of training speed or accuracy. Furthermore, we have presented a novel distributed method for exact optimization of the aggregation step for distributed ridge regression. By scaling out across 4 Titan X GPUs and using the adaptive aggregation method we were able to train on a 40GB dataset and demonstrate a $20\times$ speed-up relative to a multi-threaded distributed implementation across 4 CPU-based workers.


\section*{Acknowledgment}

The authors would like to thank Evangelos Eleftheriou, IBM Research - Zurich for his support of this work and Martin Jaggi, EPFL for useful discussions regarding distributed learning algorithms.



%\bibliographystyle{IEEEtran}
\bibliographystyle{IEEEtran}
\bibliography{IEEEabrv,ParLearning}

\end{document}